\documentclass[pdflatex,sn-mathphys]{sn-jnl}

\jyear{2021}%
\theoremstyle{thmstyleone}

\theoremstyle{thmstyletwo}%
\theoremstyle{thmstylethree}%
\usepackage{url}
\usepackage{soul}

\usepackage{array}
\usepackage{booktabs}
\usepackage{makecell}
\usepackage{longtable}
\usepackage{bbding}
\usepackage{subfigure}
\usepackage[figuresright]{rotating}
\usepackage{comment}
\usepackage{multicol}

\newcommand{\myref}[1]{Eq.(\ref{#1})}
\newcommand{\myfontsize}{\fontsize{6}{7.2}\selectfont}

\begin{document}

\title[Article Title]{Deep Learning for Iris Recognition: A Review}

\author[1]{Yimin Yin}
\equalcont{These authors contributed equally to this work.}
\author[1]{Siliang He}
\equalcont{These authors contributed equally to this work.}
\author[2]{Renye Zhang}
\author[3]{Hongli Chang}
\author[4]{Xu Han}
\author*[5]{Jinghua Zhang}\email{zhangjingh@foxmail.com}

\affil[1]{School of Mathematics and Statistics, Hunan First Normal University, Changsha, 410205, Hunan, China}
\affil[2]{School of Computer Science, Hunan First Normal University, Changsha, 410205, China}
\affil[3]{School of Information Science and Engineering, Southeast University, Nanjing, 210096, China}
\affil[4]{College of Science, National University of Defense Technology, Changsha, 410073, Hunan, China}
\affil*[5]{College of Intelligence Science and Technology, National University of Defense Technology, Changsha, 410073, Hunan, China}
%%==================================%%
%% sample for unstructured abstract %%
%%==================================%%

\abstract{Iris recognition is a secure biometric technology known for its stability and privacy. With no two irises being identical and little change throughout a person's lifetime, iris recognition is considered more reliable and less susceptible to external factors than other biometric recognition methods. Unlike traditional machine learning-based iris recognition methods, deep learning technology does not rely on feature engineering and boasts excellent performance. This paper collects 120 relevant papers to summarize the development of iris recognition based on deep learning. We first introduce the background of iris recognition and the motivation and contribution of this survey. Then, we present the common datasets widely used in iris recognition. After that, we summarize the key tasks involved in the process of iris recognition based on deep learning technology, including identification, segmentation, presentation attack detection, and localization. Finally, we discuss the challenges and potential development of iris recognition. This review provides a comprehensive sight of the research of iris recognition based on deep learning.}

\keywords{Iris recognition, Iris datasets, Deep learning, Convolutional neural networks}

\maketitle

\section{Introduction}\label{sec1}
Biological patterns, such as the face, voice, fingerprint, iris, and finger vein, have replaced traditional methods like keys, passwords, and access cards as the most reliable and effective human identification. Biological patterns, including face, voice, fingerprint, iris, and finger vein, are widely used for personal identification. Existing studies~\cite{1,2} have proven that among the aforementioned biological patterns, the iris pattern is the most accurate and secure form for personal identification due to its significant advantages: \textbf{(a) Unique:} there are not any iris having the same physical characteristic as others, even if they come from the same person or identical twins; \textbf{(b) Stability:} the iris is formed during childhood, and it generally maintains unchangeable physical characteristics throughout life; \textbf{(c) Informative:} the iris has rich texture information such as spots, stripes, filaments and coronas~\cite{2}. \textbf{(d) Safety:} Since the iris is located in a circular area under the surface of the eye between the black pupil and the white sclera, it is rarely disturbed by external factors. As a result, it is difficult to forge the iris pattern; \textbf{(e) Contactless:} \emph{Iris Recognition} (IR) is more hygienic than biometrics that requires contact, such as fingerprint recognition~\cite{3}. Due to these advantages, IR has been widely used in identification~\cite{4}.

\begin{figure}[htbp]
    \centering
    \includegraphics[width = 1.0\textwidth]{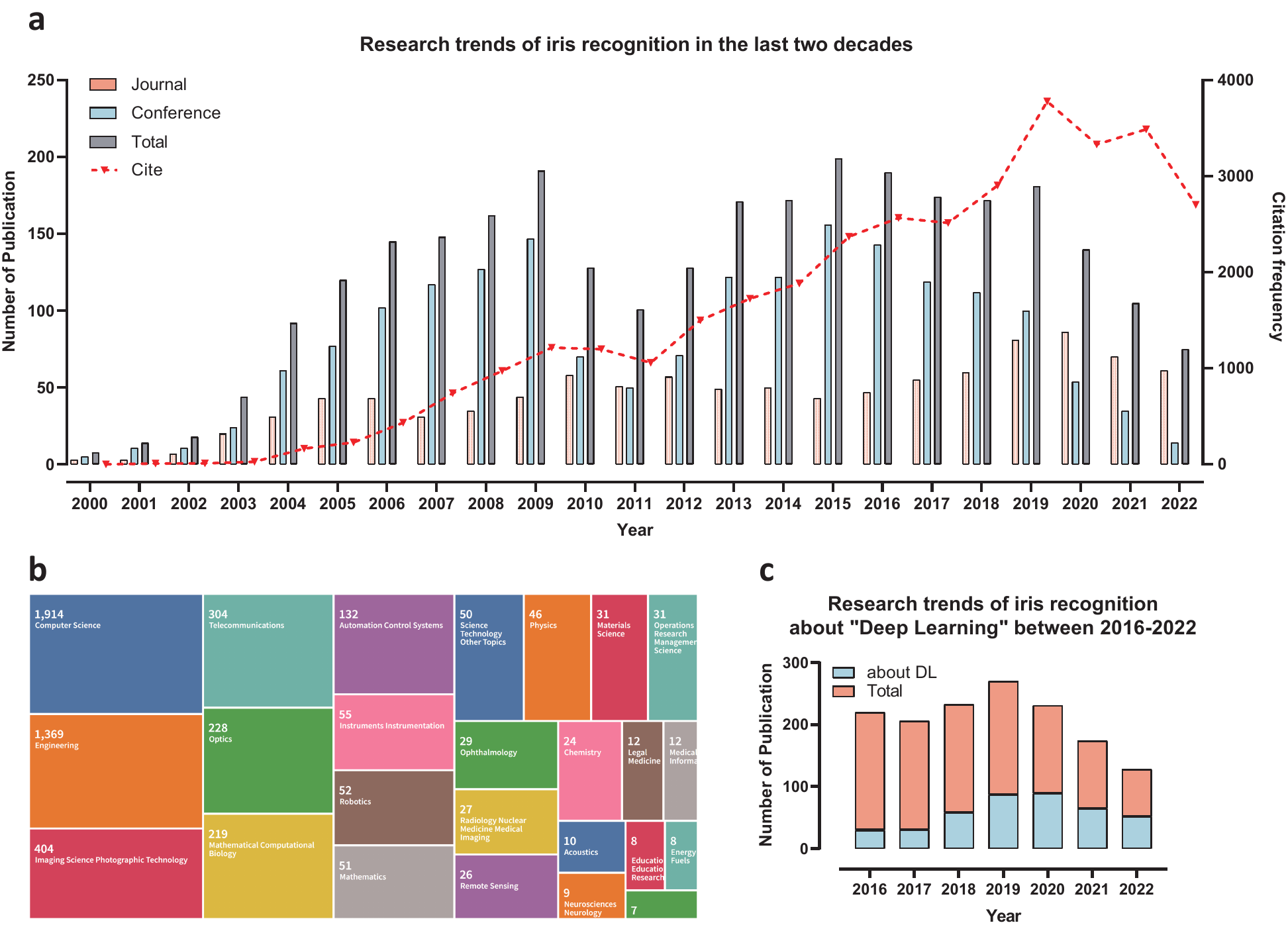}
    \caption{\textbf{(a)} Research trends of IR in the last two decades. The bar chart shows the number of papers published on the topic of \textit{IR} in the corresponding year, including conference papers and journal papers. The figure shows that academic research on IR has entered two culminations in 2008 and 2016. In the last two decades, the number of IR publications and citations has generally shown an upward trend. \textbf{(b)} The natural disciplines of IR technology. The numbers in the rectangle represent the number of related papers. \textbf{(c)} Analysis of trends in applying deep learning in IR from 2016 to 2022 based on the number of published papers. Data from Web of Science.}
    \label{iristrend}
\end{figure}

In 1993, John Daugman proposed the first automatic IR system~\cite{87}. The traditional process in most IR works first segments the iris region. Then, the desired features are extracted from these regions. Generally, most of these features are handcrafted. Due to the employment of manual features, traditional methods face some bottlenecks. For instance, they usually require a lot of pre-processing and parameter-tuning operations to handle a specific dataset~\cite{5}. Besides, many manual features are designed based on expert domain knowledge, which may hinder the mining of high-level features and result in limited performance. Since then, researchers have been seeking a more effective solution to this challenge. In 2012, AlexNet~\cite{89} overwhelmingly performed in the ImageNet competition. This is because of the superior learning capability of deep learning techniques and the large data sets available for training. Henceforth, The deep learning technique represented by \emph{Convolution Neural Network} (CNN) dominates computer vision research. Meanwhile, many IR methods based on deep learning are proposed. These methods perform better than traditional methods since deep learning can automatically learn abstract and effective feature representations from the image data.
Benefiting from the efficiency and superior performance of deep learning technology, the application of deep learning has been popular in the IR field in recent years. The trend can be found in Fig.~\ref{iristrend}. Researchers have increasingly focused on applying deep learning technology in IR, and the deep learning-related publications in the overall IR field have been continuously increasing.

IR is significant in biometrics, and deep learning currently dominates the IR task. However, the existing reviews need a comprehensive perspective. Therefore, we present the first comprehensive review of IR based on deep learning. We summarize 120 papers related to IR based on deep learning from 2016 to 2022, covering identification, segmentation, \emph{Presentation Attack Detection} (PAD), localization, image enhancement, and other tasks. 
The papers are collected from mainstream datasets or search engines, including IEEE, Springer, Elsevier, ACM, MDPI, World of Scientific, and Google Scholar. We use “IR” AND (“deep learning” OR “deep neural networks” OR “CNN”) as searching keywords.

The contributions of this paper are as follows.
\begin{itemize}
\item To the best of our knowledge, this is the first comprehensive survey summarizing the application of deep learning to IR. For this survey, we discuss 120 relevant papers. In addition, we analyse the characteristics of existing reviews and pointed out their limitations. We satisfy the usage frequency of all public iris image datasets in publications, then demonstrate a few of the most representative datasets in detail.
\item We provide a comprehensive summary for IR based on deep learning according to the various image analysis tasks. These tasks include identification, segmentation, PAD, and localization. Firstly, we present the identification tasks in non-end-to-end and end-to-end. After We summarize the iris segmentation tasks based on different network structures, they are mainly \emph{Fully Convolutional Network} (FCN) and U-Net. In the PAD task of IR, our discussion similarly observed the above perspective and divided typical neural networks and novel neural networks. In addition, the localization tasks and other tasks are also included in our survey.
\item We summarize the main challenges of applying deep learning on IR by analyzing existing papers. Meanwhile, we propose some potential development directions for the future of IR. These viewpoints provide some inspiration for other researchers.
\end{itemize}

\begin{figure}[htbp!]
     \centering
     \includegraphics[width =1.0\textwidth]{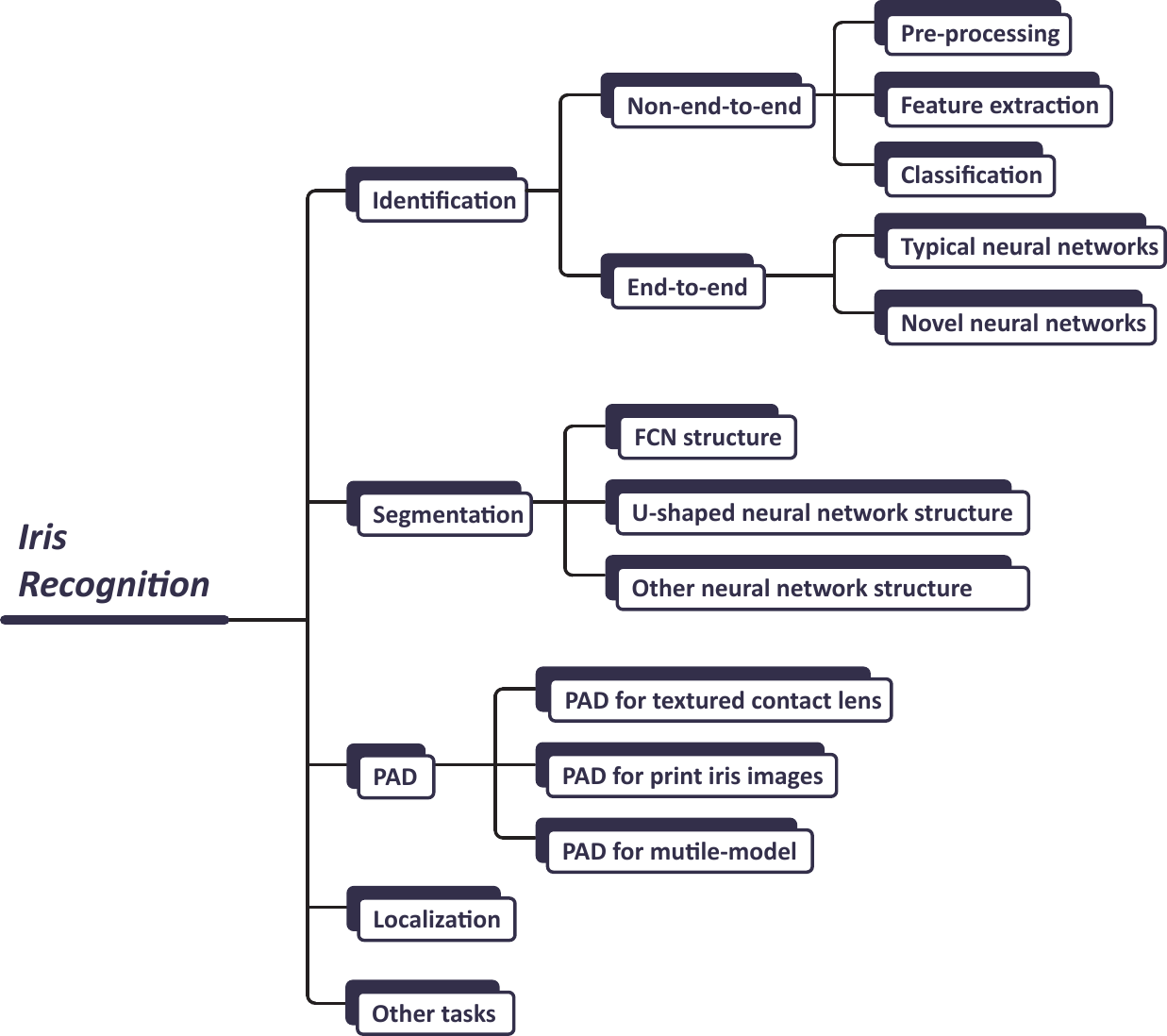}
     \caption{Overview of the related-IR tasks based on deep learning in our survey. For a comprehensive explanation of deep learning-based IR, we divided the related literature into four main tasks: identification, segmentation, PAD, and localization. This framework illustrates the further divisions in each task.}
     \label{F101}
 \end{figure}

 The structure of our review is provided in Fig.~\ref{F101}. In Section~\ref{sec2}, we analyze the characters of existing reviews and point out the limitations. Besides we also provide an overview of widely used IR datasets in this section. According to different tasks of IR, we introduce the identification task from non-end-to-end and end-to-end viewpoints in Section~\ref{sec3}. Then, we present the segmentation task of the iris based on different network structures in Section~\ref{sec4}.
 Section~\ref{sec5}, \ref{sec6}, and \ref{sec7} summarise the tasks of PAD, localization, and other tasks, respectively.
Section~\ref{sec8} presents the challenges and potential directions according to the summary of the papers mentioned above.
Section~\ref{sec9} conducts a conclusion for our review.

\section{Related work}\label{sec2}

\subsection{Compared with existing reviews}
Since IR is such a high-performance and high-penetration technology, some studies have summarized IR. However, these reviews have different limitations. In the following part, we evaluate these reviews and point out how our work differs from theirs. \cite{18,27,21} summarize the basic IR process according to the traditional biometric recognition workflow, including image acquisition, pre-processing, image segmentation, feature extraction, and classification. In~\cite{18, 27}, in addition to these above steps, image normalization is also introduced. Meanwhile, these reviews introduce machine learning-based approaches instead of focusing on deep learning methods, which cannot provide a comprehensive insight into the current deep learning-based mainstream. Specifically, \cite{18} mentions neural network techniques in the feature extraction and classification phases, and \cite{27} briefly summarizes the application of CNNs in the iris image segmentation and feature extraction phases, but there is not a comprehensive and systematic summary of deep learning techniques. Additionally, reviews~\cite{27,21} lack the summary of influential public IR datasets. Differently, our review offers a comprehensive perspective. Firstly, our review is the first comprehensive overview of the application of deep learning techniques in IR. First, our review provides a detailed overview of currently popular datasets. Then we analyze identification, segmentation, PAD, and localization tasks in terms of different image analysis tasks in IR. Finally, we illustrate the current challenges and potential directions for IR.

%\cite{32}  focuses on the analysis of IR articles conducted in a timeline. Most of the iris datasets used are also presented.

%In contrast,~\cite{15,17,20} are reviews of some of the more specific aspects of IR research, such as~\cite{15} discussing current iris segmentation methods. As can be seen from the literature surveyed in this total, many techniques have been proposed to implement iris segmentation based on edge details. The article discusses these segmentation methods according to Edge based techniques, Histogram and thresholding techniques, Clustering techniques and Contour evolution methods. Segmentation using clustering, thresholding and deformable models all yielded relatively good results. Contour evolution methods based on level sets are under development. However, no single edge-based method addresses all the challenges encountered in non-ideal iris datasets. \cite{17} reviews various approaches to IR that rely on neural networks within 2011-2019. The review proposes the use of neural networks for IR, obtaining many effective algorithms and good results. In addition, the review summarises the background of many different fields of iris image recognition techniques and compares these techniques. 

\begin{table}[htbp]
    \centering
    \caption{Summary of the related reviews according to different aspects. \Checkmark represents the corresponding review summarize or discuss the relevant content, \XSolidBrush represents no summary. \textbf{Dataset} represents the introduction of public datasets about iris images. \textbf{Deep Learning} represents the summary of the application of deep learning techniques in IR tasks. \textbf{Challenges} represents the summary of challenges in IR. \textbf{Future Directions} represents the exploration of future development directions in the field of IR.}
    \label{tab1}
    \myfontsize 
    \begin{tabular}{cccccc}
        \toprule
        \textbf{Year} & \textbf{Reference} & \textbf{Dataset} & \textbf{Deep Learning} & \textbf{Challenges} & \textbf{Potential Directions} \\
        \hline
        \specialrule{0em}{4pt}{4pt}
        2019 & \cite{18} & \Checkmark &\XSolidBrush & \Checkmark &\Checkmark \\
        \specialrule{0em}{4pt}{4pt}
        2020 & \cite{21} & \XSolidBrush &\XSolidBrush & \XSolidBrush &\Checkmark \\
        \specialrule{0em}{4pt}{4pt}
        2020 & \cite{27} & \XSolidBrush &\XSolidBrush & \Checkmark &\Checkmark \\
        \specialrule{0em}{4pt}{4pt}
       {2022} & ours & \Checkmark &\Checkmark& \Checkmark &\Checkmark \\
        \specialrule{0em}{4pt}{4pt}
        \hline
    \end{tabular}
\end{table}

\subsection{Datasets}
\begin{figure}[htbp]
    \centering
    \includegraphics[width = 1.0\textwidth]{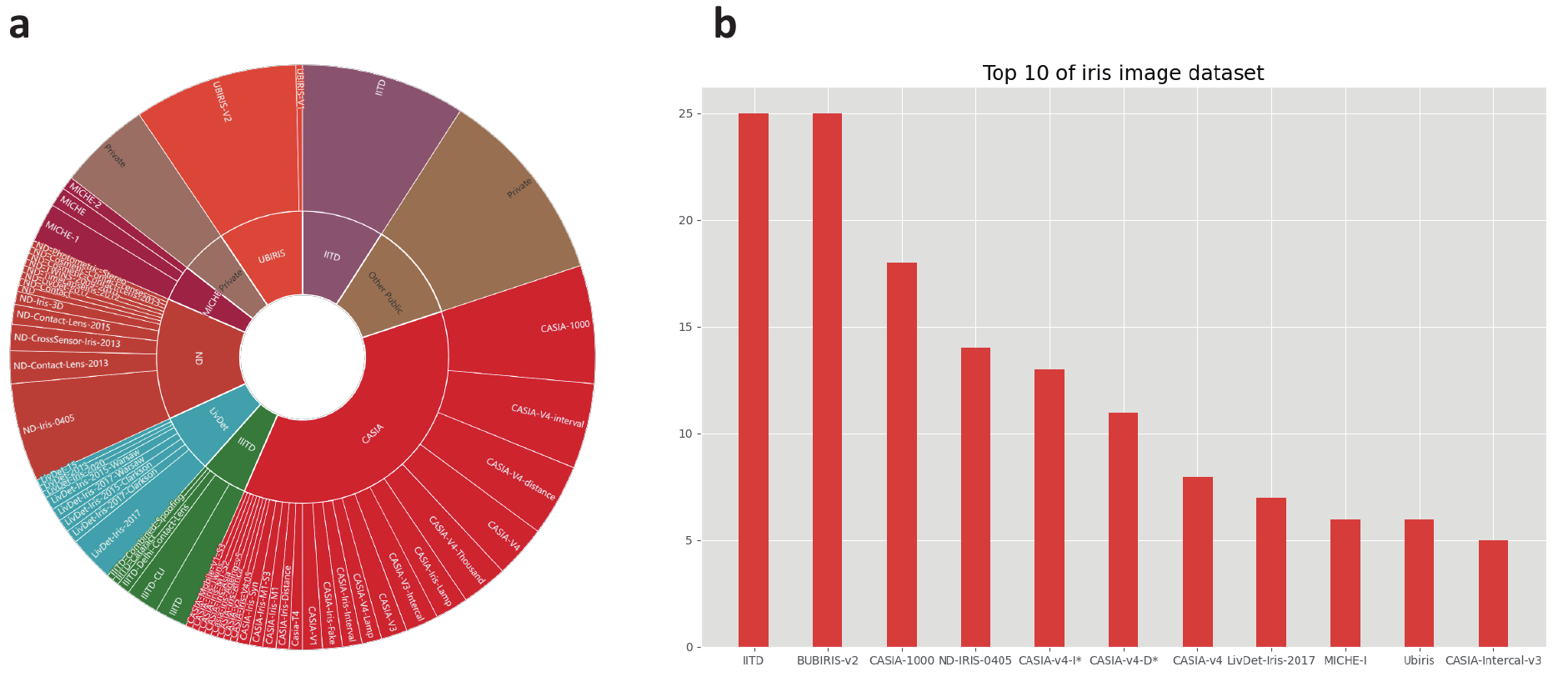}
    \caption{The public dataset of iris images is widely used in deep learning-based IR tasks, according to the findings of this paper. On the left are the statistics of the datasets in the literature covered in this paper. The right is one of the top ten frequency datasets, so this section is introduced based on the right graph.}
    \label{irisf}
\end{figure}

\begin{figure}[htbp]
    \centering
    \includegraphics[width = 1.0\textwidth]{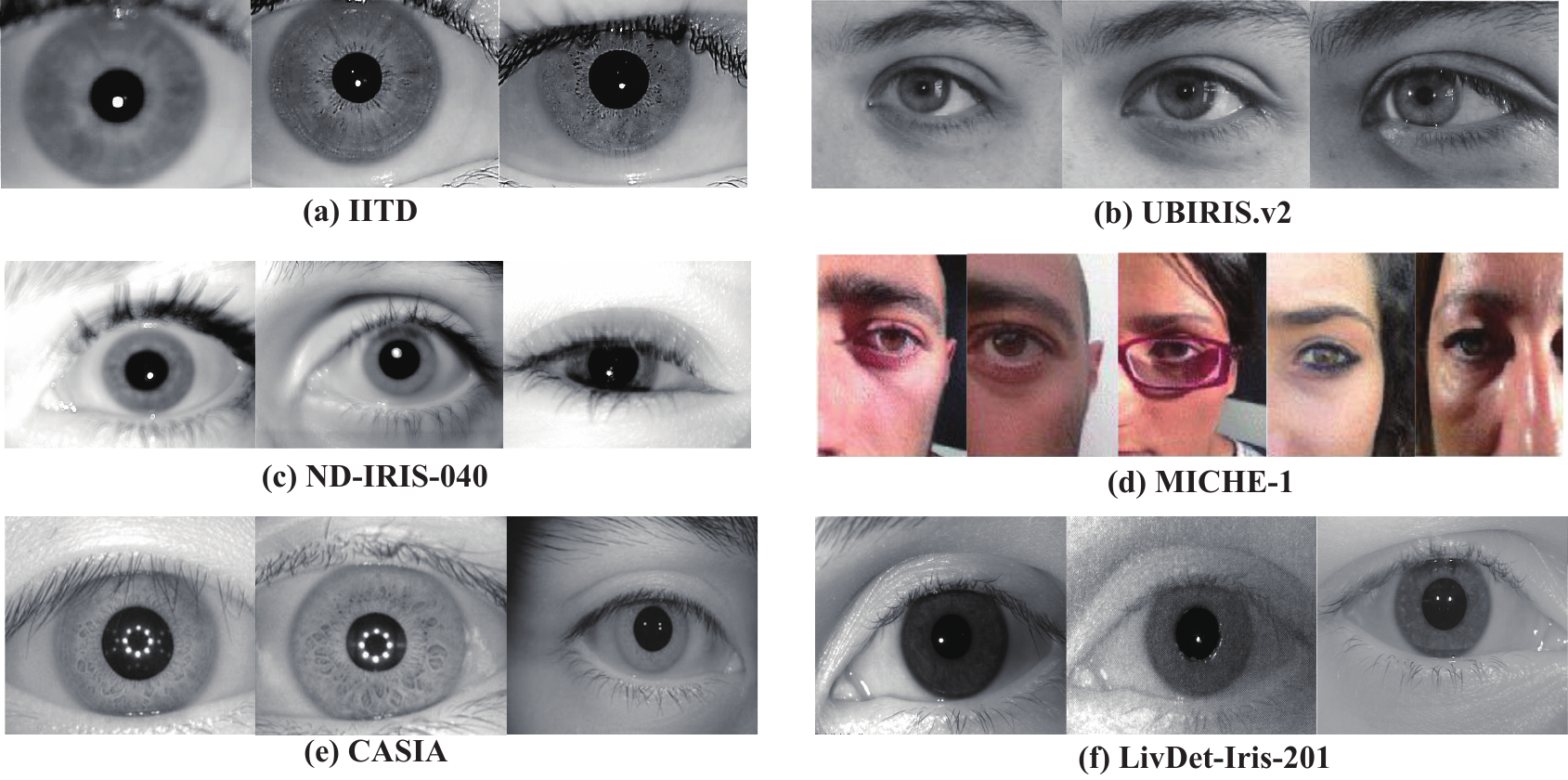}
    \caption{Some examples are obtained from the public datasets of iris images.}
    \label{database}
\end{figure}

As deep learning occupies an essential position in the research field of \emph{Artificial Intelligence} (AI), it is also widely used in IR tasks. deep learning is a data-driven learning paradigm that aims to learn effective features based on abundant training data to perform the analysis task. Therefore, the dataset plays an essential role in the development of deep learning. In this section, we introduce the commonly used datasets in IR. By investigating relevant papers, we conduct a statistic to summarize the usage frequency of various IR datasets. The details are demonstrated in Fig.~\ref{irisf}. It can be found that the popular datasets for IR mainly include IITD~\cite{95}, UBIRIS.v2~\cite{90}, ND-IRIS-0405~\cite{91}, MICHE-I~\cite{92}, CASIA-V4-Interval~\cite{94}, CASIA-V4-Distance~\cite{94}, CASIA-V4-Thousand~\cite{94}, CASIA-Iris-Lamp~\cite{94}, Clarkson Dataset~\cite{93}, Warsaw Dataset~\cite{93}, Notre Dame Dataset~\cite{93}, and IIITD-WVU Dataset~\cite{93}. The fundamental information of these datasets is provided as follows. Some examples of these datasets are shown in Fig.~\ref{database}.

\textbf{IITD} is provided by the IIT Delhi, New Delhi, India and includes a total of 2240 images collected from 224 different participants, which contain 176 males and 48 females. The age of the participants in this dataset range from 14 to 55. For each participant, five left, and five right iris images are collected. The IITD is captured using three different cameras: JIRIS, JPC1000, and complementary digital metal–oxide–semiconductor cameras. The resolution of these images is 320 $\times$ 240 pixels, and all the images are in bitmap format. Note that all these images were acquired in an indoor environment.

\textbf{UBIRIS} is an iris dataset released in 2004, and it aims to simulate imaging conditions with fewer constraints. As the initial version was criticised, the UBIRIS.v2 is built as a multi-session dataset captured under various realistic noise factors generated by unconstrained imaging conditions. The images in the first session were taken at a far distance between the camera and the participant. The second session was imaging when the participant was in a state of motion. The third session was construed under high dynamic lighting conditions. Finally, the UBIRIS.v2 dataset acquires 11102 iris images from 261 different participants with a resolution of 800 $\times$ 600 pixels. The camera used here is Canon EOS 5D.

\textbf{ND-IRIS-0405} contains 64980 images corresponding to 712 unique irises from 356 subjects. Among these subjects, 250 of the subjects are Caucasian, 82 are Asian, and the rest subjects are other races. The age range of these subjects is from 18 to 75 years old. There are 158 females and 198 males in these subjects. This dataset contains NIR images of resolution 640 $\times$ 480 pixels. These images are taken in a controlled environment with little variation.

\textbf{MICHE-I} contains 3191 VIS images captured from 92 subjects under uncontrolled settings using three mobile devices: iPhone 5, Galaxy Samsung IV, and Galaxy Tablet II (1262, 1297 and 632 images, respectively). The images have different resolutions of 1536 $\times$ 2048, 2320 $\times$ 4128, and 640 $\times$ 480 pixels, respectively. The main characteristics of MICHE-I are the wide and diverse collection population and using different mobile devices for iris collection. These images in the dataset are approaching real-world distributions.

\textbf{CASIA-V4-Interval} contains 2639 NIR iris images from 249 subjects with extremely clear iris texture details and resolution of 320 $\times$ 280 pixels, acquired in an indoor environment. These images are captured with a self-developed iris camera. To make the iris images have suitable luminous flux, the developers of this dataset sets up a circular array of cameras to capture the iris images. Benefiting from this approach, these captured images have outstanding quality with extremely clear iris texture details.

\textbf{CASIA-V4-Distance} contains 2,567 images from 142 subjects. These images were obtained using near-infrared imaging techniques. Subjects were located three meters away from the camera. The resolution of the iris images is 2,352 $\times$ 1,728 pixels. Each sample in the dataset contains the upper part of the face, which includes irises and other facial details. These details, such as skin lines, can be used for multi-modal biometric information fusion. This is a publicly available long-range and high-quality iris and facial dataset.

\textbf{CASIA-V4-Thousand} contains 20,000 NIR images from 1,000 subjects, collected in an indoor environment with different illumination settings and a resolution of 640$\times$ 480 pixels. The main source of variation in this dataset is the specular reflection of the glasses. The dataset was collected using the IKEMB-100 camera manufactured by IrisKing.

\textbf{CASIA-Iris-Lamp} contains 16,212 images from 411 subjects. The resolution of the iris images is 640$\times$ 480 pixels. Images for this dataset were acquired using a handheld iris sensor made by OKI. The acquisition environment was a room with one light turned on or off, producing a nonlinear distortion caused by changes in visible illumination. The elastic deformation of the iris texture caused by pupil dilation and contraction under different lighting conditions is one of the most common and challenging problems in IR. Therefore, CASIA-Iris-Lamp can be used to study the problem of nonlinear normalization and robust iris feature representation of the iris.
 
\textbf{Clarkson Dataset} for LivDet-Iris 2017 is collected at Clarkson University using an LG Iris Access EOU2200 camera for capturing the irises. Clarkson's training set included 2,469 real images from 25 subjects, 1,122 images from five people wearing 15 types of patterned contact lenses, and 1,346 printed images from 13 subjects. Each image is 640$\times$480 pixels. The testing set included 1,485 real images from 25 subjects, 908 printed images, and 765 from seven people wearing patterned contact lenses.

\textbf{Warsaw Dataset} includes images of real irises and corresponding images of paper prints. It includes 1,844 images of 322 different iris acquisitions and 2,669 images of the corresponding paper prints. The Iris Guard AD 100 sensor acquired all real and spoofed samples. Each print was made with a hole in the pupil position to produce the true reflection of the cornea expected by the sensor. The resolution of the samples is 640$\times$480 pixels.

\textbf{Notre Dame Dataset} consists of a total of 2700 images. The dataset is divided into a training set and a test set. The training set consists of 600 real iris images (without any contact lenses) and 600 images of eyes of people wearing \emph{Textured Contact Lenses} (TCLs) made by Vapor, University College London, and ClearLab. The test set consists of 1800 real iris images and 1800 images of the eyes of people wearing TCLs. Of the real iris images, 900 of these are collected from the same location as the training set, and the other 900 are different. Of the images of eyes of people wearing textured contact lenses, 900 are produced by Vaporware, University College London, and ClearLab, and 900 images are produced by Coopor and J\&J. The resolution of all iris images is 640$\times$ 480 pixels, and all images are acquired by LG 4000 and AD 100 sensors.
 
\textbf{IIITD-WVU Dataset} is a hybrid of two datasets. The IIITD-WVU training set consists of 2,250 real images and 1,000 iris images from subjects wearing textured contact lenses from the IIIT-Delhi dataset. These images were acquired in a controlled environment. Also, 3,000 printed attack images were selected from the IIITD iris spoof dataset. The testing set of the IITD-WVU dataset consists of 4,209 images, including 702 real iris images, 701 iris images with textured contact lenses, 1,404 printed iris images, and 1,402 iris images with textured contact lenses. The testing set of the IIITD-WVU is a novel multi-session iris PAD dataset. The images in this dataset were acquired using the IriShield MK2120U mobile iris sensor at two different places, indoors (controlled lighting) and outdoors (different environmental conditions).

\begin{sidewaystable}[htbp]
    \centering
    \caption{A critical summary of datasets information that is widely used in IR. \textbf{Subjects} and \textbf{Images} represnts the total number of subjects and samples.}
    \renewcommand{\arraystretch}{1.5}
    \label{tab:my_label}
    \myfontsize 
    \begin{tabular}{ccccllllc}
        \toprule
        \multicolumn{3}{c}{\textbf{Dataset}} & \textbf{Spectrum} & \textbf{Subjects} & \textbf{Images} & \textbf{Resolution} & \textbf{Format} & \textbf{URL} \\
        \hline
        \multicolumn{3}{c}{IITD} & NIR & 224 & 2240 & $320\times 240$ & bmp & \cite{95} \\
        
        \multicolumn{3}{c}{UBIRIS.v2} & VIS & 261 & 11102 & $800\times 600$ & tiff & \cite{90} \\
       
       \multicolumn{3}{c}{ND-IRIS-0405} & NIR & 356 & 64980 & $640\times 480$ & jpeg & \cite{91}\\
       
       \multicolumn{3}{c}{MICHE-1} & VIS & 30 & 240 & $960\times 1280$ & jpg & \cite{92} \\
       \hline
        \multirow{4}{*}{CASIA} &\multicolumn{2}{c}{CASIA-V4-Interval} & NIR & 249 & 2639 & $320\times 280$ & jpeg & \multirow{4}{*}{\cite{94}} \\
       
       ~ & \multicolumn{2}{c}{CASI-V4-Distance} & NIR & 142 & 2567 & $2,352\times 1,728$ & jpeg &  ~ \\
       
       ~ & \multicolumn{2}{c}{CASIA-V4-Thousand} & NIR & 1000 & 20000 & $640\times 480$ & jpeg & ~ \\
        
       ~ & \multicolumn{2}{c}{CASIA-Iris-Lamp} & - & 411 & 16212 & $640\times 480$ & jpeg & ~ \\ 
       \hline
        \multirow{9}{*}{LivDet-Iris-2017} & \multirow{2}{*}{Clarkson} & training set & NIR & 43 & 4937 & $640\times 480$ & PNG & \multirow{10}{*}{\cite{93}}\\
        ~ & ~ & testing set & NIR & 68 & 4066 & $640 \times 480$ & PNG & ~ \\
        \cline{2-3}
        ~ & \multirow{2}{*}{Warsaw} & training set & NIR & 322 & 4513  & $640\times 480$ & PNG & ~\\
        ~ & ~ & testing set & NIR & 50 & 2990 & $640\times 480$ & PNG & ~\\
        \cline{2-3}
        ~ & \multirow{2}{*}{Notre Dame} & training set & NIR & - & 1200 & $640\times 480$ & PNG & ~\\
        ~ & ~ & testing set & NIR & - & 3600 & $640\times 480$& PNG & ~ \\
        \cline{2-3}
        ~ & \multirow{2}{*}{IIITD-WVU} & training set & NIR & - & 6250 & $640 \times 480$ & PNG & ~\\
        ~ & ~ & testing set &NIR & - & 4209 & $640 \times 480$ & PNG & ~ \\ 
        \bottomrule
    \end{tabular}
\end{sidewaystable}

\section{Iris identification}\label{sec3}
Although the dominant research trend in deep learning is to use end-to-end models to process data, there are still many studies in iris identification using the non-end-to-end approach. Therefore, we discuss the iris identification task in two parts, non-end-to-end and end-to-end.

\subsection{Non-end-to-end approaches}
\begin{figure}[htbp]
    \centering
    \includegraphics[width = 1.0\textwidth]{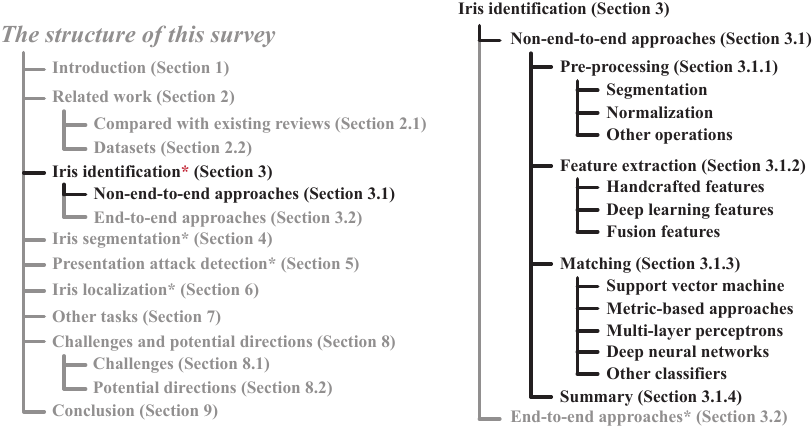}
    \caption{The structure of \textbf{\textit{Non-end-to-end approaches}} in \textbf{\textit{Iris identification}}.}
    \label{structure_identification_ne}
\end{figure}

In some non-end-to-end deep learning paradigms, the iris identification process usually consists of three steps: pre-processing, feature extraction, and matching. In this section, we follow the above principles in our discussion and discuss each of the three steps in the non-end-to-end approach. Each method we present in this section is fragmented and cannot do iris recognition on its own, but these methods are integral steps in the non-end-to-end iris recognition process. The structure of this section is shown in Fig.~\ref{structure_identification_ne}.

\subsubsection{Pre-processing}
In many non-end-to-end deep learning approaches, image pre-processing is required to improve the iris image quality to facilitate the following analysis so that it can obtain a more accurate recognition rate. Generally, the widely used pre-processing operations in the iris identification task include iris segmentation, normalization, highlights removal, and image enhancement.

\paragraph{Segmentation}
The image segmentation in the iris identification task aims to segment the iris region in people's eyes accurately.
Traditionally, classical threshold-based segmentation is used to perform this task, which aims to seek a suitable threshold to binaries the image based on the lower grey value of the pupil~\cite{76}. However, a single threshold value makes the segmentation method lack generalization ability. In order to make the iris images obtained in different environments well segmented, the Hough transform model has become a more commonly used method in iris segmentation in recent years~\cite{35,75,71,76,77,73}.

The Hough transform is an algorithm with a cumulative voting mechanism. It finds the boundary parameters by judging whether each boundary point in the image space satisfies the possible trajectories constituted by the parameter space. And the target contour fitting is achieved based on the boundary parameters. For instance, \cite{71,77,73} also utilize the Canny operator and Hough transform to detect the outer boundary of the iris. These studies applied the Hough transform model directly to \emph{Region of Interest} (RoI) to localize the iris efficiently. Differently, \cite{76} employs the binarization method to locate the inner boundary points and the Hough transform to discover the inner boundary circles for iris segmentation.

In addition to the Hough transform, Gaussian filters, a linear smoothing filter suitable for removing Gaussian noise, are also commonly used for iris segmentation. For example, \cite{11} detects rounded edges using a Gaussian filter and iterative search, and \cite{78} uses the Gaussian filter to find weak and robust edges by calculating local maxima and setting different thresholds for locating soft and firm edges, respectively. This method is less susceptible to Gaussian white noise and more robust.

There are also some iris segmentation methods in pre-processing that use deep learning techniques. \cite{67,71} uses Mask R-CNN to detect the iris location on the entire image and segment the iris region based on the previous localization results.
\cite{68} uses the FCN to perform the segmentation on pre-processing. FCN can work without fixing the original image size and inputting the whole iris eye image without overlapping regions for double computation.

\paragraph{Normalization}

Normalization is a technique used to convert data with different measurement scales into a common scale, allowing for meaningful comparison and analysis. This process has the added benefit of minimizing the impact of outliers by deflating the scale of outliers, which can greatly influence the distribution and statistical properties of the data. Normalization is a common pre-processing step used in iris recognition based on end-to-end modelling to improve the accuracy and stability of results.

\begin{figure}[h]
     \centering
     \includegraphics[width =0.5\textwidth]{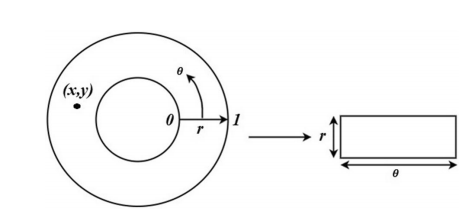}
     \caption{Daugman’s rubber sheet model~\cite{2}}
     \label{Daugman’s rubber sheet model}
 \end{figure}

Many iris normalization processes employ the Daugman rubber sheet model~\cite{87,35,64,81,11,80,68,77}. As shown in Fig.~\ref{Daugman’s rubber sheet model}, the Daugman rubber sheet model can transform the iris image from Cartesian coordinates to polar coordinates. Specifically, Daugman’s model takes each point $(x,y)$ within the iris region to a pair of normalized non-concentric polar coordinates $(r,\theta )$ where $r$ is on the gap $\left [ 0, 1 \right ]$ and $ \theta $ is the angle on the interval $[0, 2 \pi]$. The mathematical definition of the iris region mapping is shown in~\myref{Eq401}:
\begin{equation}    \label{Eq401}
\left\{\begin{matrix}
 I(x(r,\theta )) & \rightarrow & I(r,\theta )\\
 x(r,\theta ) & = & (1 - r){x_p}(\theta )r{x_l}(\theta )\\
 y(r,\theta ) & = & (1 - r){y_p}(\theta )r{y_l}(\theta )
\end{matrix}\right.,
\end{equation}
where $I(x, y)$ is the intensity value of $(x, y)$ in the image of the iris region. The parameters ${x_p}$, ${x_l}$, ${y_p}$, and ${y_l}$ are the coordinates of the pupil and iris boundaries along the $\theta$ direction. The advantage of using this model is that circular regions can be easily converted to rectangular shapes~\cite{68}. Besides, the Daugman rubber sheet model used by~\cite{77} considers the size and pupil dilation. The iris RoI is segmented and then normalized, thus reducing the effect of iris noise and improving recognition efficiency.

\paragraph{Other operations}
In addition to the abovementioned operations, removing highlighted areas and image enhancement are also commonly used in iris identification.

\begin{figure}[htbp]
  \centering
  \subfigure{
    \includegraphics[scale = 0.9]{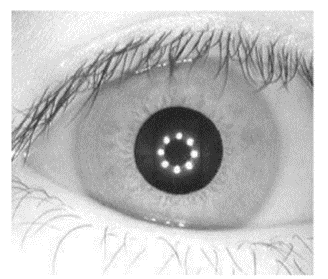}}
  \hspace{0.15in} 
  \subfigure{
    \includegraphics[scale = 0.9]{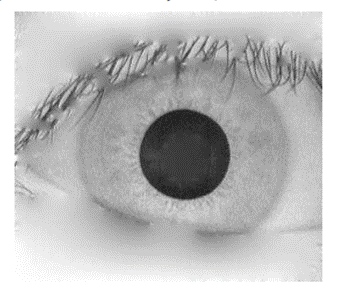}}
  \caption{(a) CASIA Dataset Sample Iris (Resolution $320 \times 280$). (b) Removal of specular highlight via linear interpolation.}
  \label{Thisislabel} 
\end{figure}

\cite{78} removes highlights from the original iris image during pre-processing to reduce the detrimental effects on the detection process. This removal operation is conducted by a two-dimensional linear interpolation of the bright spots of the detected light caused by reflections. Fig.~\ref{Thisislabel} illustrates the effect of this removal operation: The \textit{left} is the original iris image, and the \textit{right} is the processed image by removal of specular highlights via linear interpolation.

Some studies also focus on image enhancement methods~\cite{75,76,86,71}. \cite{75,76} use the histogram equalization method. The cumulative distribution function enhances the contrast of the image without changing the original brightness space of the image. This method enhances the normalized image, improves the image quality, and makes the iris texture more prominent. Besides, \cite{86} enhances the normalized region using a defogging algorithm with a dark channel prior. Additionally, \cite{71} uses black hat processing, median processing, and gamma correction to improve the quality of the input image.

\subsubsection{Feature extraction}
In the past, iris recognition required the extraction of various handcrafted features to complete the recognition process. The design of these handcrafted features usually relied on biological and computer experts. Since recent years, with the booming of deep learning in various visual tasks, the application of \emph{Deep Neural Networks} (DNN) for feature extraction has received much attention in iris recognition~\cite{35}. However, according to our survey, many research efforts still use handcrafted methods based on the domain's prior experiences to extract iris features. Therefore, this section summarizes not only the feature extraction methods based on deep learning but also sorts out the handcrafted methods for extracting iris features.

\paragraph{Handcrafted features}

Traditional handcrafted feature extraction methods often require tedious steps. For instance, \cite{77} uses a fusion method including scattering transform, Tetrolet transform, local gradient pattern, and local optimal orientation pattern to extract iris features. Then \emph{Deep Belief Network} DBN is used for classification. Although this scheme eventually obtains an accuracy of 97.9\%, the model inference steps are tedious. Traditional machine learning methods for extracting features like this are expensive. Deep learning methods can automatically extract high-level semantic features from massive data. To illustrate the difference, \cite{81} compares some typical handcrafted feature extraction methods with CNN feature extraction methods. The handcrafted schemes include Log-Gabor, Contourlet transform, and local gradient autocorrelation. The experimental results show that the CNN feature extraction methods outperform the handcrafted methods mentioned above.

\paragraph{Deep learning features}

With the continuous development of deep learning techniques, using DNNs to extract features has become a trend in iris recognition. For instance, \cite{11} compared the feature extraction capabilities of different popular deep CNNs, including AlexNet, VGGNet, GoogLeNet, ResNet, and DenseNet.
These deep CNNs have tens or even hundreds of layers with millions of parameters, and they are good at capturing and encoding complex features from iris images. In the feature extraction process of CNNs, each layer of the CNN models the visual content of the image at a different level. The earlier layers retain coarser information, while the later layers encode finer and more abstract features. This approach uses the output of each layer as feature descriptors and reports the corresponding recognition accuracy. The experimental results show that among the five different CNNs, the sixth layer of DenseNet performs the best in the LG2200 dataset with a recognition accuracy of 98.7\%. The fifth layer of DenseNet achieves a recognition accuracy of 98.8\% in the CASIA-Iris-Thousand dataset. These preliminary results show that CNNs effectively extract useful visual features from iris images, eliminating laborious and expensive manual feature engineering tasks.

Based on our analysis of related literature, it is common to employ AlexNet, DenseNet, and ResNet as independent feature extractors for non-end-to-end iris matching tasks. \cite{35} uses pre-trained AlexNet to extract features and SVM to perform classification. \cite{72} proposes a \emph{Multi-Instance Cancelable Iris System} (MICBTDL). MICBTDL uses the operational triple loss of AlexNet training for feature extraction and stores the feature vectors as cancelable templates. The network of~\cite{37} is an adaptation of DenseNet-161 and is called \emph{Densely Connected Contact Lens Detection Network} (DCLNet). Unlike the original DenseNet-161~\cite{43}, DCLNet retains only two dense blocks for feature extraction, effectively reducing the complexity of the network. \cite{73} uses the ResNet-101~\cite{98} and DenseNet-201~\cite{99} architectures. ResNet introduces the concept of residual connectivity. DenseNet utilizes the concept of densely connected layers. These two compact network architectures improve computational and storage efficiency. The model achieves an accuracy of 96\%. \cite{102} uses a ResNet-50 without a fully connected layer is used to capture the initial feature map. Afterwards, global and multiscale features were generated using the worldwide mean and multiscale ensembles, respectively.

In addition, some studies utilize self-designed network structures to extract iris features. In~\cite{67}, the pre-processed iris images are fed into FeatNet to extract features and perform matching. FeatNet collects different levels of semantic information from different convolutional layers and adjusts them to the same feature dimension. A convolutional layer is then used to fuse these features to generate a single-channel feature map that preserves the correspondence with the original input in the feature space. \cite{8} is an extension of~\cite{67}. The model proposed by~\cite{8} is called UniNet. It consists of two sub-networks, FeatNet and MaskNet. Both subnetworks are generated based on FCN. UniNet is used for feature extraction. This network can perform pixel-to-pixel mapping by merging upsampling layers. Also, an extended tristate loss function specifically for iris features is designed in this study.

Instead of using NIR images alone, \cite{124} uses both NIR and VIS iris images for matching. Their pre-processed images are fed into two separate generators, using discriminators that discriminate the spectral domain and contrast loss to optimize the generator parameters so that the generators can align different spectral iris image features from the same identity. Finally, the generator-generated feature maps are fed into a convolutional neural network for further feature extraction and classification.

\paragraph{Fusion of handcrafted and deep learning features}
In addition to the simple use of handcrafted and deep learning features, some works combine both deep learning and handcrafted features to obtain better performance. They can complement each other to some extent. For instance, \cite{64} explores complementary features to improve the accuracy of iris recognition for mobile devices. First, optimized \emph{Ordinal Measures} (OMs) features are extracted to encode local iris textures. Then, CNNs are used to automatically learn paired features to measure the correlation between two irises. Finally, the selected OMs features and the paired features extracted by the CNN are fused to improve the recognition performance further.

% The features obtained by fusing the handcrafted features and the features extracted by deep learning are complementary. It makes the iris-matching results more effective. \cite{64} explores complementary features to improve the accuracy of iris recognition for mobile devices. First, optimized \emph{Ordinal Measures}(OMs) features are extracted to encode local iris textures. Then, CNNs are used to automatically learn paired features to measure the correlation between two irises. Finally, the selected OMs features and the paired features known by the CNN are fused at the score level.

\subsubsection{Matching}
Matching is the final step in iris recognition, which aims to match the features extracted from iris images with their corresponding registered user. In the non-end-to-end approaches, matching is usually performed by classifiers independent of the feature extractor. In this section, according to the frequency of the classifiers that appear in non-end-to-end methods, we introduce SVM, DNNs, KELM, and MLP in detail.

\paragraph{Support vector machine}
SVM is popular as a machine learning algorithm for matching in iris identification~\cite{38}. Its basic model is a linear classifier with a maximum margin defined on the feature space, and the maximum interval makes it different from traditional models. SVM is usually combined with various models for feature extraction to perform iris identification tasks~\cite{7,11,33,34,35,36,37,41}. For example, \cite{35,37} combine the simple SVM with the feature extractor based on DNNs for iris recognition. These works show a good recognition rate on relevant datasets. In addition to simply using a DNN for feature extraction, \cite{36,41} perform dimensionality reduction using \emph{Principal Component Analysis} (PCA)~\cite{42} on the features extracted by the network to reduce the complexity. PCA is a usual dimensionality reduction method that removes redundant information from the original data and retains important features. The models used for feature extraction in these two studies are ResNet-50 and VGG-16, respectively. The experimental results show that these two studies obtain the ACC of 96.41\% on the CASIA-Iris-Thousand dataset and the \emph{Recognition Rate} (RR) of 99.4\% on the IITD dataset. To further verify the classification ability of SVM, \cite{34} compared the classification performance of KNN and SVM. These two algorithms classify the features extracted by CNN. The experimental results show that the classification performance of SVM is better than that of KNN. In contrast, KNN cannot handle large datasets and is more sensitive to the noise introduced due to image enhancement.

\paragraph{Metric-based approaches}
Metric learning measures the distance between feature vectors, which accurately compares and classifies similar and dissimilar feature vectors. It is widely used in non-end-to-end approaches to iris identification. Based on our survey, Euclidean distance, cosine similarity, and \emph{Hamming Distance} (HD) are the more widely used methods, so this section focuses on these metrics.

\textbf{Euclidean Distance} is a basic measure of distance in vector space, which determines the linear distance between feature vectors. This metric is simple and widely used for various classification tasks, including IR. \cite{102} uses the Euclidean distance of global features and the spatial reconstruction distance of multi-scale features is computed to perform matching the feature vectors.

\textbf{Cosine similarity} measures the difference between two feature vectors according to their cosine of the angle. Compared with the normal Euclidean distance, the cosine distance focuses more on the difference between two vectors in the direction. \cite{80} also uses cosine similarity for matching, but the difference is that this model uses segmentation and normalization techniques in the pre-processing stage.
In~\cite{73}, matching is performed using cosine similarity. The model, a CNN trained on a closed set, is first used as a feature extractor. Then a separate binary classifier is trained for each identity. This step allows the model to classify using a simple similarity metric, speeding up the matching phase and simplifying the registration process. The experimental results show a recognition accuracy of 97.3\% in closed recognition and 98.5\% in open set recognition.

\textbf{HD} compares the number of different pixels between two feature vectors of the same length. The fewer the different pixels, the more the two feature vectors are from the same iris. The HD is much easier to calculate than the previous cosine similarity. In~\cite{64}, the features extracted by manual and deep learning methods are used, respectively. Moreover, it uses different matching ways. In the manual extraction method, the matching process is done by calculating the HD of OMs features to measure the correlation between feature pairs. In another method, matching is done automatically by the fully connected layer of the CNN and the Softmax function. \cite{85} proposes to incorporate a supervised discrete hashing scheme to reduce the size of the iris template significantly. It uses the features extracted from CNN to perform iris matching using HD. \cite{8} also uses HD for classification.
Similarly, \cite{65} designed a lightweight CNN architecture for a small iris dataset. AlexNet is used for feature extraction, and HD is used to compute image similarity. The experiments were performed on ND-IRIS-0405 with an \emph{Equal Error Rate} (EER) = 0\%.

\paragraph{Multi-layer perceptrons}
A \emph{Multi-layer Perceptron} (MLP) is a type of feedforward neural network that consists of shallow layers of interconnected nodes capable of learning non-linear relationships between inputs and outputs. Compared to DNNs, MLP has a smaller number of parameters, making them easier to train and faster to converge. 
\cite{82,83} both use MLP in the matching stage. \cite{82} uses MLPs and CNNs as classifiers to compare the effectiveness of extracted features generated by CNNs and hand-crafted features, respectively. When using the MLP, the hand-crafted features (local binary patterns) produce better prediction accuracy than the features extracted by the CNN.
The MLP trained by~\cite{83} uses softmax cross-entropy loss to achieve binary classification decisions. The periocular-assisted iris identification In less constrained imaging environments, iris images present different regions of effective iris pixels. The difference in the adequate number of available iris pixels can be used to dynamically enhance the periocular information obtained from the iris images.
\cite{72} uses MLP as the matching module. Experiments were conducted on the iris datasets of IITD and MMU to verify the effectiveness of the framework of MICBTDL.

\paragraph{Deep neural networks}
With the rise of deep learning, DNNs perform the matching task excellently. According to our survey, DBN is heavily used in non-end-to-end iris identification matching tasks. Other typical DNN structures are generally used to perform end-to-end iris identification tasks, so in this section, we only discuss work related to DBNs.

A DBN with a modified feedforward neural network based on a backpropagation algorithm is used in~\cite{78} to classify contour-based iris features. The system achieves a high classification rate even at a low signal-to-noise ratio. The classification accuracy of the method is 99.92\%.
\cite{75} uses an adaptive Gabor filter determined by particle swarm optimization and binary particle swarm optimization. This filter adapts to the frequency band that is richest in iris information. The obtained iris Gabor codes are then put into a DBN, which is then used to detect potential learning features of the iris in a data-driven manner. The Gabor codes of all registrants and their corresponding labels are then assigned to the DBN model for training. Finally, the DBN algorithm is used to implement feature extraction and matching of the iris.
\cite{76} proposes an effective stacked \emph{Convolution Deep Belief Network} (CDBN) combined with DBN for multi-source heterogeneous iris identification. The stacked CDBNs-DBN model is designed to extract and identify similar feature structures between MSH iris textures. The role of CDBNs is to extract the similar texture feature structure of MSH iris images. DBNs are classifiers. Using DBNs as classifiers can reduce the reconstruction error through the negative feedback mechanism of the self-encoder. \cite{77} uses DBNs based on the steepest gradient for matching. The maximum accuracy of this classification method is 97.9\%.

\paragraph{Other classifiers}

In addition to the above classifiers, other classifiers also show good performance in the matching phase of IR. 

\cite{66} proposes a matching method for iris features using a capsule network. These iris features were extracted using pre-trained typical neural networks, including GoogLeNet, Iris-Dense, and VGG-16. The matching uses a capsule network based on a dynamic routing algorithm. This algorithm takes into account the direction and length information of the feature vectors and can effectively estimate the pose information of the object. The experimental results show that the matching method achieves 99.42\% accuracy in the JluV4 dataset when using Iris-Dense for feature extraction.

The identification of~\cite{84} is a matter of variable selection and regularization. The model uses sparse linear regression techniques to infer matching probabilities. The strategy is robust to outlier match scores, a significant source of error in biometrics.
\cite{86} uses a multi-layer similar convolutional structure to reduce computational complexity. This structure reduces the dimensionality of the iris texture information and obtains the primary texture information of the iris. Finally, a collaborative representation is used to extract and classify the iris textures.

\subsubsection{Summary}
This section describes the non-end-to-end iris identification approach. In the pre-processing part, image segmentation is almost an essential step. Because iris features are more delicate than other biometric features, extracting the iris region from the whole image helps the subsequent model to extract iris features. In addition, the Hough transform is also widely used, enabling the iris region to be segmented even in the presence of a complex background. DNN shows the same powerful performance as various visual tasks in the feature extraction part. The performance of handcrafted methods is equally satisfactory, but their models are complex, and inference costs are high for iris identification tasks. In the matching phase, SVM is popular as a classical classification method because it can effectively match iris features even with small data samples. Many studies use metric-based methods to match, using Cosine Similarity and HD. The experimental results of these studies show that the design of HD is more suitable for the matching task of iris features. In addition to the classical DNN structure, some studies have also used capsule networks for the matching task, demonstrating the superiority of dynamic routing algorithms. We provide a summary table for these papers in Tab.~\ref{tab3}.

\begin{sidewaystable}[htbp]
    \centering
    \caption{Summary of the IR based on a non-end-to-end approach. DRS represents the Daugman rubber sheet model. CUDR represents correct U/D recognition. BOTH represents Both handcrafted extracted features and features extracted by deep learning are used. CRR represents the correct recognition rate. }
    \label{tab3}
    \myfontsize 
    \begin{tabular}{cclllcc}
        \toprule
        \multirow{2}*{\textbf{Year}} & \multirow{2}*{\textbf{Reference}} & \multicolumn{3}{c}{\textbf{Method}} & \multirow{2}*{\textbf{Dataset}} & \multirow{2}*{\textbf{Results}} \\
        \cline{3-5}
        ~ & ~ & \textbf{Per-processing} & \textbf{Feature extraction} & \textbf{Matching} & ~ & ~ \\
        \hline
         \specialrule{0em}{4pt}{4pt}
        2016 & \cite{78} & \makecell[l]{Gaussian filter \\ Image enhancement} & Radius vector function& DBN & \makecell[l]{CASIA-Iris-V4} & \makecell[l]{ACC=99.92\% }\\
         \specialrule{0em}{4pt}{4pt}
        2016 & \cite{64} & DRS & BOTH & HD & \makecell[l]{Private}  & \makecell[l]{EER=0.48\%} \\
        \specialrule{0em}{4pt}{4pt}
         2016 & \cite{81} &  DRS & BOTH & KNN, SVM, and KELM & \makecell[l]{CASIA-v4}  & \makecell[l]{ACC=98.6\% }\\
          \specialrule{0em}{4pt}{4pt}
        2016 & \cite{41} &  $\backslash$  & \makecell[l]{VGG-16 \\PCA} &SVM & \makecell[l]{IITD}   & \makecell[l]{ACC=99.4\%}  \\
          \specialrule{0em}{4pt}{4pt}
        2017 & \cite{11} & \makecell[l]{Gaussian filter \\ DRS} &  off-the-shelf CNNs & SVM & \makecell[l]{ND-CrossSensor-2013 \\ CASIA-Iris-Thousand} & \makecell[l]{DenseNet-ACC = 98.8\% \\ AlexNet-ACC = 98.5\%} \\
          \specialrule{0em}{4pt}{4pt}
        2017 & \cite{75} & \makecell[l]{Hough transform \\ Canny edge detection}  & Gabor & DBN & \makecell[l]{CASIA-V4-Interval\\ CASIA-V4-Lamp} & \makecell[l]{CRR = 99.998\% \\ CRR = 99.904\%} \\
          \specialrule{0em}{4pt}{4pt}
        2017 & \cite{8} & \cite{202}  & FCN & HD & \makecell[l]{ND-IRIS-0405 \\ CASIA-V4-Distance\\IITD\\WVU Non-ideal}  & \makecell[l]{EER=0.99\% \\ EER=3.85\%\\EER=0.73\%\\EER=2.28\%} \\
          \specialrule{0em}{4pt}{4pt}
        2017 & \cite{33} & $\backslash$  & handcrafted feature & SVM & \makecell[l]{Private} & \makecell[l]{CUDR=98.1\%} \\
        \specialrule{0em}{4pt}{4pt}
        2017 & \cite{65} & $\backslash$  & AlexNet  & HD & \makecell[l]{ND-IRIS-0405}  &  \makecell[l]{EER=0.00\%}\\
          \specialrule{0em}{4pt}{4pt}
        2017 & \cite{82} & $\backslash$  & handcrafted feature  & MLP or CNN & \makecell[l]{"Gender from Iris"} & \makecell[l]{ACC=66$\pm$2.7\%}\\
        \specialrule{0em}{4pt}{4pt}
        \bottomrule
    \end{tabular}
\end{sidewaystable}

\begin{sidewaystable}[htbp]
    \centering
    \caption{Continued: Summary of the IR based on a non-end-to-end approach. \textbf{DRS } represents the Daugman rubber sheet model. \textbf{IIITD} represents the IIIT-Delhi Contact lens iris database.}
     \myfontsize 
  \begin{tabular}{cclllcc}
        \toprule
        \multirow{2}*{\textbf{Year}} & \multirow{2}*{\textbf{Reference}} & \multicolumn{3}{c}{\textbf{Method}} & \multirow{2}*{\textbf{Dataset}} & \multirow{2}*{\textbf{Results}} \\
        \cline{3-5}
        ~ & ~ & \textbf{Per-processing} & \textbf{Feature extraction} & \textbf{Matching} & ~ & ~ \\
        \hline
      
        \specialrule{0em}{4pt}{4pt}
        2018 & \cite{35} & \makecell[l]{Hough transform \\ Canny edge detection} & pre-trained Alex-Net & SVM & \makecell[l]{IIT Delhi Iris  \\ CASIA-Iris-V1\\CASIA-Iris-Thousand\\CASIA-Iris-Interval} & \makecell[l]{ACC=100\%\\ACC=98.3\%\\ACC=98\%\\ACC=89\%}\\
        \specialrule{0em}{4pt}{4pt}
        2018 & \cite{80} & DRS & CNN & cosine similarity & \makecell[l]{UBIRIS } & \makecell[l]{EER=13.98$\pm$0.55\% }\\
        \specialrule{0em}{4pt}{4pt}
         2018 & \cite{84} &  $\backslash$ & CNN & Sparse linear regression & \makecell[l]{CASIA-Iris-Thousand \\ UBIRIS.v2}& \makecell[l]{EER=0.32\% \\ EER=2.73\%} \\
        \specialrule{0em}{4pt}{4pt}
         2019 & \cite{86} &  Image enhancement & CNN & Collaborative representation & \makecell[l]{MICHE-I }& \makecell[l]{RR=97\% }\\
        \specialrule{0em}{4pt}{4pt}
        2019 & \cite{67} & Mask R-CNN & FeatNet & FeatNet &\makecell[l]{ND-IRIS-0405 \\ CASIA-V4-Distance\\IITD\\WVU Non-ideal}  & \makecell[l]{EER=1.12\% \\ EER=4.07\%\\EER=0.76\%\\EER=2.20\%} \\
        \specialrule{0em}{4pt}{4pt}
        2019 & \cite{37} & $\backslash$ & DenseNet-161 & SVM &\makecell[l]{IIITD }  & CCR=99.10\% \\
         \specialrule{0em}{4pt}{4pt}
        2019 & \cite{68} & \makecell[l]{FCN \\ DRS} & MCNN & MCNN   & \makecell[l]{UBIRIS.v2\\ LG2200\\CASIA-Iris-Thousand}  & \makecell[l]{ACC = 99.41\% \\ ACC = 93.17\% \\ ACC = 95.63\%} \\
         \specialrule{0em}{4pt}{4pt}
        2019 & \cite{36} & $\backslash$ & Fine-tuned ResNet-50 & SVM & \makecell[l]{CASIA-Iris-Thousand} & \makecell[l]{ACC = 99.41\%} \\ 
        \specialrule{0em}{4pt}{4pt}
         2019 & \cite{66} & $\backslash$ & \makecell[l]{GoogLeNet\\Iris-Dense \\VGG16} & DRDL & \makecell[l]{JluV3.1 iris \\JluV4\\ CASIA-V4-Lamp} & \makecell[l]{ACC = 99.37\%\\ACC = 99.42\%\\ACC =93.87\%} \\
        \specialrule{0em}{4pt}{4pt}
        \bottomrule
    \end{tabular}
\end{sidewaystable}

\begin{sidewaystable}[htbp]
    \centering
    \caption{Continued: Summary of the IR based on a non-end-to-end approach. TH represents Threshold segmentation and Hough transform. DRS represents the Daugman rubber sheet model.}
    \myfontsize 
\begin{tabular}{cclllcc}
        \toprule
        \multirow{2}*{\textbf{Year}} & \multirow{2}*{\textbf{Reference}} & \multicolumn{3}{c}{\textbf{Method}} & \multirow{2}*{\textbf{Dataset}} & \multirow{2}*{\textbf{Results}} \\
        \cline{3-5}
        ~ & ~ & \textbf{Per-processing} & \textbf{Feature extraction} & \textbf{Matching} & ~ & ~ \\
        \hline
        
        \specialrule{0em}{4pt}{4pt}
          2019 & \cite{85} & $\backslash$ & CNN & HD & \makecell[l]{ UND 2012 cross sensor \\PolyU bi-spectral\\Cross-eye-cross-spectral } & \makecell[l]{EER = 4.50\%\\EER =5.39\%\\EER =6.34\%} \\
        \specialrule{0em}{4pt}{4pt}
             2020 & \cite{124} & $\backslash$ & U-Net &  $\backslash$ &  \makecell[l]{PolyU dataset} & \makecell[l]{EER=1.02\% } \\
        \specialrule{0em}{4pt}{4pt}
         2020 & \cite{83} & $\backslash$ & UniNet & MLP(Dynamic Fusion) & \makecell[l]{CASIA-Moblie-V1-S3\\CASIA-V4-\\ Q-FIRE } & \makecell[l]{EER= 0.73\%\\EER = 2.29\%\\EER=3.87\%} \\
        \specialrule{0em}{4pt}{4pt}
        2021 & \cite{76} & \makecell[l]{TH} & 2D Gabor & DBN & \makecell[l]{IITD} & \makecell[l]{ACC = 99.3\%} \\
        \specialrule{0em}{4pt}{4pt}
        2021 & \cite{71} &\makecell[l]{Hough transform \\ Mask R-CNN } & $\backslash$&  GoogLeNet & \makecell[l]{CASIA-Iris-Thousand} &\makecell[l]{ACC = 99.14\%}  \\
        \specialrule{0em}{4pt}{4pt}
        2021 & \cite{77} & \makecell[l]{Hough transform \\DRS} & The handcrafted features & DBN & \makecell[l]{CASIA Iris} & \makecell[l]{ACC =97.9\% }\\
        \specialrule{0em}{4pt}{4pt}
        2021 & \cite{73} & \makecell[l]{Hough transform \\DRS} & ResNet-101 and DenseNet-201 & cosine similarity & \makecell[l]{CASIA Iris} & \makecell[l]{ACC =97.9\% }\\
            \specialrule{0em}{4pt}{4pt}
        2021 & \cite{102}  & $\backslash$ & ResNet-50 & \makecell[l]{Euclidean distance\\Spatial reconstruction distance}& \makecell[l]{ND-IRIS-0405\\CASIA-V4-Distance\\ CASIA-M1-S3\\UBIRIS.v2} &\makecell[l]{EER=0.75\%\\ EER=2.17\%\\  EER=1.63\%\\ EER=5.58\%} \\
        \specialrule{0em}{4pt}{4pt}
        2022 & \cite{72} & \makecell[l]{Hough transform \\ Canny} &  CNN trained using triplet loss & ANN & \makecell[l]{IITD\\ MMU } & \makecell[l]{EER = 0.06\%\\EER=0.03\%} \\
        \specialrule{0em}{4pt}{4pt}
         2022 & \cite{34} & Image cropping &  CNN & \makecell[l]{KNN\\SoftMax\\SVM} & \makecell[l]{IITD} & \makecell[l]{ACC = 86\%\\ACC=95.16\%\\ACC=97.8\%} \\
        \specialrule{0em}{4pt}{4pt}
        \bottomrule
    \end{tabular}
\end{sidewaystable}

\subsection{End-to-end approaches}
\begin{figure}
    \centering
    \includegraphics[width = 1.0\textwidth]{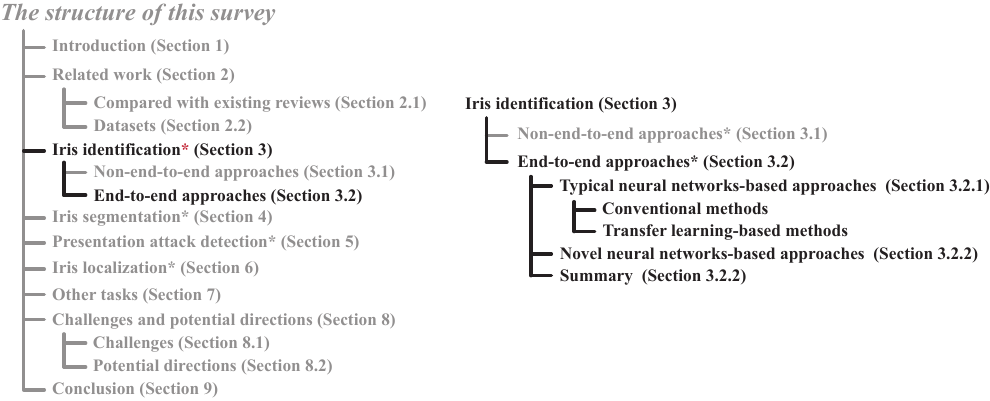}
    \caption{The structure of \textit{\textbf{End-to-end approaches}} in \textit{\textbf{Iris identification}}.}
    \label{structure_identification_e}
\end{figure}

Many typical DNN structures have achieved excellent success in various vision tasks, especially those networks that have obtained superiority in ILSVRC. Therefore, many studies use these typical networks to recognize iris images in end-to-end IR identification tasks, pre-trained on ImageNet and trained from scratch on iris datasets. In addition, some studies have designed specialized end-to-end architectures for the IR identification task, showing satisfactory performance. These networks are also discussed in this section. The structure of this section is shown in Fig.~\ref{structure_identification_e}.

\subsubsection{Typical neural networks-based approaches}
In this section, we discuss some studies that use typical neural networks to perform the identification task of IR. In addition, many studies use transfer learning technology to save computational costs and training time when performing end-to-end identification tasks using typical networks. Therefore, we divide this proportion into two parts: conventional methods and transfer learning.

\paragraph{Conventional methods}
Some studies introduce residual connections when designing network structures for IR tasks.
\cite{120} uses the residuals connection from ResNet. Since residual connections allow the network to be built deeper without degrading performance. The model used in~\cite{120} is the \emph{Collaborative Convolutional Residual Network} (CCRNet). Experimental results show that CCRNet has good performance with an EER of 1.06\% and 1.21\% for ND-CrossSensor-Iris-2013 and ND-IRIS-0405. As an extension of residual connection, the DenseNet establishes a dense connection between all previous layers. Specifically, it connects all layers directly. To guarantee the feed-forward process, each layer splices the input from all previous layers and later propagates the feature map to all subsequent layers.
\cite{116} proposes an end-to-end approach for iris identification using \emph {Dense Deep Convolutional Neural Network} (DDNet), which consists of two parts, a transition layer and a dense block, using a deeper network structure. Also, this study uses LabelMe~\cite{117} to manually label eyelashes as non-iris regions, which will help analyse eyelashes' effect on the iris.

The acquired iris images are not clear in various acquisition environments. \cite{74} is dedicated to matching iris images containing noise rather than a simple matching task. An effective defence mechanism based on \emph{Discrete Wavelet Transform} (DWT) for adversarial attacks is proposed in~\cite{74}. The model reconstructs various denoised versions of the iris image based on the DWT. A deep CNN based on U-Net is used for feature extraction and classification. Experimental analysis of a benchmark iris image dataset, IITD, yielded good results with an average accuracy of 94\%.

\paragraph{Transfer learning-based methods}
Pre-training DNNs usually implement transfer learning, and using the pre-trained models can save much effort by initializing experienced parameters for the new task. Many researchers on IR tasks use DNNs pre-trained on ImageNet and then fine-tuned on iris image datasets rather than training new models~\cite{100} from scratch. These DNNs mainly include VGGNet, ResNet, GoogLeNet, and AlexNet, \textit{etc}. The knowledge learned by these DNNs on ImageNet can be efficiently transferred to IR tasks through transfer learning techniques.

AlexNet is the first deep CNN to achieve a breakthrough in image classification tasks. It is the champion of ILSVRC2012 and a landmark network in the whole field of deep learning. 
\cite{109,111} use AlexNet to perform identification tasks of IR. \cite{111} uses high-quality iris images selected from public datasets to train AlexNet, rather than all images in the dataset. The image spatial quality evaluator evaluates the quality score of the iris images, and images with high scores are involved in the training process. \cite{109} does not filter the training samples but performs some pre-processing operations on the input images. These operations include the Gaussian filter, triangular blur averaging method, and triangular blur median smoothing filter. These operations can improve the signal-to-noise ratio by blurring the regions outside the boundaries. The experimental results illustrate that higher-quality images and adequate pre-processing can significantly improve the identification performance of AlexNet for iris images. The above approaches successfully apply the AlexNet to IR, but the parameters of AlexNet are too large. To perform identification tasks on IR using the more lightweight network, \cite{113} uses a pre-trained SqueezeNet method for iris identification. SqueezeNet~\cite{112} is an adaptation of AlexNet, and it compresses parameters to 1/50 of AlexNet without reducing the accuracy. The approach dramatically reduces the training cost and ensures accuracy in iris identification.

VGGNet~\cite{106}, the second-place winner at ILFVRC2014, has a shallower network structure than AlexNet. It uses convolutional layers composed of small convolutional kernels instead of larger convolutional layers. On the one hand, the smaller the convolutional kernel, the fewer parameters are needed when the perceptual fields are the same. On the other hand, reducing the size of the convolutional kernels corresponds to performing more non-linear mapping, which helps to enhance the model fitting ability. \cite{107} uses the segmented iris images by Hough transform and Canny edge detection to train the VGG-16. Tests were performed on the IITD, CASIA-Iris-V1, CASIA-Iris-Thousand, and CASIA-Iris-Interval, obtaining accuracies of 100\%, 98.3\%, 95\%, and 91.6\%, respectively. \cite{186} uses the entire iris images to input VGG-16. This scheme trains on the post-mortem iris dataset and classifies both live and post-mortem irises. The CRR of this method is nearly 99\%, unlike the methods mentioned above, which use the iris alone for recognition. The biometric system of \cite{108} is a multi-model recognition model, including the iris, face, and finger vein. The network structure of this system is also a pre-trained VGG-16.
The results show that using three biometric patterns in this biometric system gave better results than using two or one biometric patterns.

AlexNet and VGGNet focused on improving performance by increasing the depth of the network. However, too-deep structures bring bottlenecks of overfitting, unacceptable computational cost, and gradient disappearance to DNN. Based on these obstacles, GoogLeNet has been designed through the Inception model. Inception is to put multiple convolution or pooling operations together in a module and design the network as these modules assemble the whole network. The Inception can autonomously select the required convolution kernel size based on the size during forward propagation. This allows the network to extract more efficient features for the same amount of computation. 
\cite{126} uses the pre-trained GoogLeNet for the identification of IR. The proposed method tests on the IITD iris dataset with an ACC of 97.51\%.
Different from normal recognition tasks, \cite{114} aims to analyze the influence of iris segmentation on iris identification. It trains Xception~\cite{115} using the segmented iris images and the entire iris images separately and then tests them on an identical test set. The experimental results show a better performance of the recognition model when the segmentation is sufficiently precise. Xception is an improvement of the Inception module, using the depth-wise separable convolution to replace the previous convolution.

Numerous experiments have shown that the deeper the network, the harder it is to train. ResNet uses residual connections to build the network to curb this phenomenon rather than a stack of convolution layers. AlexNet is the champion of ILSVRC2015; it is by far the most frequently used backbone in deep learning research. \cite{118} uses a network structure optimization strategy to adjust parameters of ResNet-18~\cite{98} to obtain better performance on identification task of IR. This scheme treats the network design process for iris identification as a constrained optimization problem, using model size and computation as learning criteria to find the optimal iris network structure with the highest recognition accuracy. The final obtained tuned network structure has higher accuracy than the original ResNet-18 with EER=1.54\% versus 1.29\% for the same case with less computation and memory requirements.

\cite{101} compares the performance of three typical networks on identifying IR, including VGG-16, ResNet-50, and GoogLeNet. The final experimental results show that the GoogLeNet achieves the best performance, a \emph {Correct Recognition Rate} (CRR) of 99.64\%. on CASIA-Iris-Distance.

\subsubsection{Novel neural networks-based approaches}
Rather than basing iris identification on existing network structures, many studies have changed the network structure of CNNs or the number of parameters to design new network architectures for iris identification tasks. The study's results demonstrate that these networks achieve good results on iris identification tasks.

CNNs have been successful on many visual tasks, as well as on IR tasks. However, some DNNs add other filters to CNNs in order to enhance feature extraction capabilities. The framework of~\cite{127} proposes a domain adaptation solution for IR, incorporating a pairwise filtering layer in the network. In the training phase, a pair of heterogeneous iris images are fed into a pairwise filtering layer to extract domain-invariant features by reducing the differences between domains. The two respective feature maps are combined into a single feature map reflecting the differences between domains, which is then fed into a subsequent convolutional layer to extract high-level semantic information further. The EER values of this research method on Q-FIRE and CASIA are 0.15\% and 0.31\%, respectively. The iris recognition technique proposed by~\cite {142} uses a local circular Gabor filter for initial feature extraction before input to the CNN to retain all directional information. This design solves the problem of traditional Gabor wavelet transform insensitivity to circular orientation versus the difficulty of neural networks extracting directional features on the circular structure of the iris. \cite{132} proposes an iris recognition method explicitly designed for post-mortem samples, thus enabling the application of iris biometrics in forensic science. This paper combines traditional Gabor wavelet-based iris coding with a DNN driven by post-mortem iris data for feature extraction, reducing the recognition error rate by one-third compared to the baseline method~\cite{178}.

In addition to filters, attention mechanisms and feature histogram methods are also applied to optimize the feature extraction process of IR. \cite{134} proposes a spatial attention feature fusion module to fuse features at different levels. Spatial attention~\cite{135,136} can encode the different positions of each importance in the feature map. The fact that iris features are local features suggests that iris features have other spatial significance in different local regions; therefore, spatial attention feature fusion is well suited for iris recognition. The spatial attention feature fusion module in the dual spatial attention network proposed in this study can learn the weights of each location and effectively fuse features at different levels. The experimental results show that the identification EER of the baseline model combined with the spatial attention feature fusion module is 0.27\%.
\cite{137} proposes ETENet, a network built using an attention-based ETEblock, which contains multiple convolutional layers and uses an attention mechanism to interact with iris features between different convolutional layers. The ETEblock enables the final extracted iris features to incorporate information from different dimensions and thus contain rich semantic information. \cite{133} proposes \emph{Histogram Iris Network} (HsIrisNet). This model mainly uses the image histogram. An image's histogram can represent the number of pixels per hue in the image. HsIrisNet uses histograms of selectively robust regions upsampled from the image to identify users. Experimental results of selecting these regions show that the technique can reduce attacks on iris biometrics. The experimental results show that the model achieves 99.72\% and 99.39\% accuracy on CASIA and IITD datasets, respectively.

Some studies have incorporated traditional machine learning methods in DNNs to make the extracted iris features with more accurate semantic information than neural networks alone. \cite{4} proposes an open-set iris identification based on deep learning techniques. Deep networks train the method to cluster the extracted iris features around the feature centres of each iris image and then build an open class feature outlier network containing distance features before mapping the extracted features to a new feature space and classifying them. The experimental results show that the method has good open-set IR performance and can effectively distinguish unknown classes of iris samples. \cite{141} proposes a network model focusing on a single interaction block. The interaction block establishes the relationship between adjacent and long-term features by computing the affinity matrix between all high-level feature pairs. Meanwhile, the interaction block is designed to be independent of the network structure and applies to different deep network structures. Experiments show that the network has an EER of 1.27\%, 0.82\% and 0.81\% on ND-IRIS-0405, CASIA-V4-Thousand and CASIA-V4-Lamp, respectively.

Optimizing the feature extraction process is important for IR tasks because it helps to improve the matching performance. However, the computational resource is limited in most IR application scenarios, so it is also necessary to design recognition models with small parameters. \cite{130} uses residual network learning with an extended convolutional kernel to optimize the training process and gather contextual information from the iris image. This approach reduces the need for downsampling and upsampling layers, and the network structure is simplified. Experimental results obtain the matching accuracies of 97.7\%, 87.5\%, and 96.1\% on ND-IRIS-0405, CASIA-V4-Distance, and WVU Non-ideal, respectively. \cite{131} proposes a lightweight CNN. The model utilizes iris features from different convention layers in this network and combines features describing the environment. Combining features relating to the environment helps reduce intra-class variation and decreases the EER.

Most studies use an iris image for identity matching, which is simple and efficient. However, some studies use both left and right iris for matching to improve accuracy. In \cite{2}, both left, and right irises are fed into the network simultaneously, their respective match scores are generated, and finally, the two scores are fused to obtain the final match. Experimental results show that the system achieves an accuracy close to 100\% on all three datasets, SDUMLA-HMT, CASIA-V3-Interval, and IITD, and identifies users in less than one second. \cite{138} also uses left and right iris images for classification. In addition, this study proposes a chaos-based encryption algorithm. This encryption algorithm combines a logic map, and a linear feedback shift register ~\cite{139} state sequence that can effectively protect the iris sample before it is sent to the server.

\subsubsection{Summary}
Many researchers have used existing typical networks for training or modification. There are some studies of novel networks. All have achieved good performance in the IR process. \cite{101} has multiple networks for comparison experiments. The selected CRR is the best performance among the multiple networks. \cite{108} results in a fusion of face, iris, and finger vein with score levels. In the~\cite{114} experiments, the best results are obtained using a model with an automatic segmentation step. To illustrate the application of DNNs on IR, we provide a summary table of related papers in Tab.~\ref{tab4}.

\begin{sidewaystable}[htbp]
    \centering
     \caption{Summary of the IR based on End-to-end approach. \textbf{RER} represents rejection rate. \textbf{WBPv1} represents Warsaw-BioBase-PostMortem- Iris v1 dataset. \textbf{CASIA-Iris-T.} represents CASIA-Iris-Thousand and SBVPI. \textbf{RER} represents rejection rate. \textbf{WBPv1} represents Warsaw-BioBase-PostMortem-Iris v1 dataset.}
    \label{tab4}
    \myfontsize 
    \begin{tabular}{cccccccccc}
        \toprule
        \textbf{Year} & \textbf{Reference}  & \textbf{Network} & \textbf{Datasets} & \textbf{Performance} & \textbf{Year} & \textbf{Reference}  & \textbf{Network} & \textbf{Datasets} & \textbf{Performance}\\
        \hline
        \specialrule{0em}{4pt}{4pt}
        2016 & \cite{127} & Custom& \makecell[l]{Q-FIRE \\ CASIA } & \makecell[l]{EER=0.15\%  \\EER=0.31\%} & 2020 & \cite{101} & GoogLeNet & \makecell[l]{CASIA-distance \\UBIRIS.v2} & \makecell[l]{CRR=99.64\%  \\CRR=98.76\%} \\
         \specialrule{0em}{4pt}{4pt}
          2018 & \cite{186}   &  VGG-16 & \makecell[l]{WBPv1} & \makecell[l]{CCR=99\%} & 2020 & \cite{108} & VGG-16 & \makecell[l]{SDUMLA-HMT } & \makecell[l]{ACC=100\%}\\
        \specialrule{0em}{4pt}{4pt}
        2018 & \cite{2} & Custom & \makecell[l]{CASIA-Iris-V3 \\IITD} & \makecell[l]{ACC=99.82\% \\ACC=99.87\%} & 2020 & \cite{111} & AlexNet & \makecell[l]{CASIA } &\makecell[l]{ ACC=100\%}\\
        \specialrule{0em}{4pt}{4pt}
         2019 & \cite{130} & Custom & \makecell[l]{ ND-IRIS-0405\\ CASIA-V4-Distance\\WVU Non-ideal} & \makecell[l]{ACC=97.7\%\\ACC=87.5\% \\ACC=96.1\%} & 2021 & \cite{138} & Custom &  \makecell[l]{Phoenix }&\makecell[l]{ACC=100\% } \\
        \specialrule{0em}{4pt}{4pt}
         2019 & \cite{131} & Custom & \makecell[l]{CASIA-Iris-M1-S2 \\ CASIA-Iris-M1-S3 \\Iris-Mobile } & \makecell[l]{EER=0.14\%\\EER=1.90\%\\EER=1.16\%} & 2020 & \cite{133} & Custom& \makecell[l]{CASIA-V3-Interval\\IITD } & \makecell[l]{CRR=99.72\%\\CRR=99.39\% }\\
        \specialrule{0em}{4pt}{4pt}
        2019 & \cite{107} &VGG-16 & \makecell[l]{IITD \\CASIA-Iris-V1\\CASIA-Iris-Thousand\\  CASIA-Iris-Interval}  & \makecell[l]{ACC=100\%\\ ACC=98.3\%\\ACC=95\%\\ACC=91.6\%} & 2021 & \cite{134} & Custom & \makecell[l]{CASIA-V4-Thousand\\CASIA-V4-Distance\\IITD } & \makecell[l]{EER=0.27\%\\EER=3.23\%\\EER=0.45\%} \\
        \specialrule{0em}{4pt}{4pt}
        2019 & \cite{109}  & AlexNet & \makecell[l]{Cogent } & \makecell[l]{ACC=86.8\%} & 2021 & \cite{137} & Custom & \makecell[l]{CASIA-V4-Lamp\\CASIA-V4-Thousand\\IITD } & \makecell[l]{EER=0.07\%\\EER=0.63\%\\EER=0.02\%}  \\
        \specialrule{0em}{4pt}{4pt}
        2019 & \cite{114}  & GoogLeNet & \makecell[l]{CASIA-Iris-Th.} &\makecell[l]{ACC=98.91\% } & 2021 & \cite{113} & AlexNet & \makecell[l]{SDUMLA-HMT \\IITD } & \makecell[l]{ACC=99.1\%\\ACC=99.7\%}\\
        \specialrule{0em}{4pt}{4pt}
        2019 & \cite{116} & DenseNet & \makecell[l]{CASIA-Iris-V3\\IITD} & \makecell[l]{CRR=99.84\%\\CRR=99.78\%} & 2022 & \cite{4} & Custom &\makecell[l]{$\backslash$} & \makecell[l]{RER=99.94\% }\\
        \specialrule{0em}{4pt}{4pt}
        2020 & \cite{118} & ResNet-18 & \makecell[l]{CASIA-Iris-Thousand\\UBIRIS.v2} & \makecell[l]{EER=1.54\%\\EER=5.98\%} & 2022 & \cite{141} & Custom &  \makecell[l]{ND-IRIS-0405\\CASIA-V4-Thousand\\CASIA-V4-Lamp} & \makecell[l]{EER=1.27\%\\EER=0.82\%\\EER=0.81\%}\\
        \specialrule{0em}{4pt}{4pt}
        2020 & \cite{120} &ResNet & \makecell[l]{ND-CrossSensor-Iris-2013\\ ND-IRIS-0405} & \makecell[l]{EER=1.06\%\\EER=1.21\%} & 2022 & \cite{126} & GoogLeNet &  \makecell[l]{IITD } & \makecell[l]{ACC=97.51\%}\\
        \specialrule{0em}{4pt}{4pt}
        2022 & \cite{74} & U-Net &  \makecell[l]{IITD } & \makecell[l]{ACC=94\% } & 2022 & \cite{142} &Custom &  \makecell[l]{CASIA-Iris-Syn \\ CASIA-Iris-Lamp} & \makecell[l]{ACC=99.70\%  \\ACC=98.62\%}  \\
        \hline
    \end{tabular}
\end{sidewaystable}

\section{Iris segmentation}\label{sec4}
\begin{figure}[htbp]
    \centering
    \includegraphics[width = 1.0\textwidth]{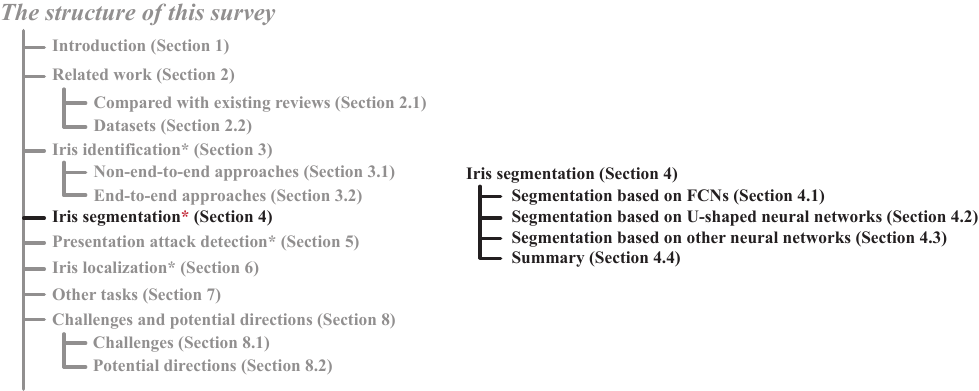}
    \caption{The structure of \textbf{\textit{Iris segmentation}}.}
    \label{structure_identification_s}
\end{figure}

Iris segmentation is a crucial step in iris recognition. Feeding a segmented iris image into a recognition model can often achieve better performance than directly using the entire iris image. Traditional manual iris segmentation methods are computationally conveniently complex and require extensive expert experience. To perform iris segmentation more easily and accurately, many studies have focused on segmenting iris images using deep learning methods. FCN is a classical network for image segmentation tasks, and U-structured networks are commonly used segmentation network structures. Therefore, this section categorises these studies into three parts: FCN-based iris segmentation, U-shaped network-based segmentation, and other networks. The structure of this section is shown in Fig.~\ref{structure_identification_s}.

\subsection{Segmentation based on fully conventional neural networks}
\cite{52} presents the \emph{Fully Convolutional Neural Network} (FCN) structure, which was initially proposed for semantic object segmentation tasks. Still, the network structure accepts arbitrarily sized inputs and produces outputs of corresponding sizes through efficient inference and learning. It has been shown that this network structure improves the segmentation performance of iris images. 

Some studies use the original FCN as the backbone to optimize the model to better adapt to iris image segmentation tasks. \cite{54} improves the up-sampling method in the original FCN by combining the current feature channels with the corresponding feature channels, enabling down-sampling layers to require a high-resolution iris image. \cite{55} adds dilated convolution to the FCN, which allows the network to extract more globalized features from the iris image than regular convolution. Inspired by DenseNet, \cite{60} adds the dense block to the network, and each dense block contains three convolution layers. This dense block composites the feature values from the bottom layer to the top layer, which can effectively suppress overfitting and gradient vanishing. In contrast to the above methods, \cite{62} does not change the network structure but sets the segmentation targets to multi-classes instead of just iris and non-iris regions. The training process of semantic segmentation is more complex than the segmentation based on two classes but captures more semantic information from the original iris image.

To extract more detailed features from the iris image, \cite{53} proposes two different iris segmentation models that are called \emph{Hierarchical Convolutional Neural Network} (HCNN) and \emph{Multiscale Fully Convolutional Network} (MFCN), respectively. Both models can automatically locate iris locations instead of relying on manual production. In HCNN, the same iris image is resized into three patches of different sizes and fed to each of the three CNN channels to learn features. Finally, a fully connected layer labels every pixel on the image. However, the computational volume of this model is complex since some patches are repeatedly computed, and the fixed size input size limits the receptive fields of convolution layers. To overcome these problems, this study ten proposes MFCN without the fully connected layer, where images can be input at arbitrary sizes. This design allows the MFCN to capture rough and detailed features in the iris image accurately. The experimental results show that MFCN has more noise immunity than HCNN, and MFCN obtains the EER of 0.29\% on UBIRIS.v2 and 0.20\% on CASIA-V4-Distance.

The full convolution encoder-decoder structure is widely used in image segmentation. \cite{55, 56, 58, 63} use an encoder-decoder network to perform the iris image segmentation task. The encoder and decoder of~\cite{55} are both 44 layers, and~\cite{55} introduce the dropout to inhibit overfitting and Monte-Carlo sampling to output the model uncertainty for each class. The experimental results show that the network with the new strategy has better segmentation performance than the original encoder-decoder structure.
\cite{56} uses the same encoder-decoder network as~\cite{55} for iris image segmentation. Besides, to solve the problem of the lack of labelled data in the IR task, this study designs two domain adaptation methods based on linear mapping and non-linear mapping, respectively. These domain adaptation methods transfer the data distribution from the source domain to the target domain to generate a new sample distribution, train the encoder-decoder network in the new sample distribution, and finally, use the trained model to segment the iris images on the target domain. The F score is the harmonic mean of the fraction of relevant instances among the retrieved instances and the fraction of relevant instances retrieved over the total relevant instances. The experimental results show that this method obtains an F score of 0.925, 0.937, and 0.951 on CASIA-V5-Aging, CASIA-V4-Interval, and IITD, respectively.
In~\cite{58}, the encoder structure consists of the first 13 layers of VGG-16, and the feature map generated by the fifth pooling layer is used as input to the decoder. The decoder uses three transposed convolutional layers for upsampling and adds skip layers to extract high-resolution features. Experiments on various public datasets show that the \emph{Average Segmentation Error} (ASE) of the network is satisfactory.
\cite{63} simulates the training process of a GAN rather than just an encoder-decoder-based segmentation network. The framework consists of an encoder-decoder network and a generator comprising a CNN-based discriminator, with an additional jump connection between the encoder and decoder to fuse shallow and deep features. In the training of this model, ground truth images and segmented images generated by the encoder-decoder are fed to the discriminator. Then the losses are back-propagated to the encoder-decoder to optimize this network. This training method further improves the segmentation performance of the model. In the encoder, the three remaining blocks are densely connected to improve the feature extraction capability. Also, multi-supervised training of multi-scale images is used in this study to reduce the effect of image size. Experimental results show that the framework achieves \emph{mean Intersection over Union} (mIoU) of 95.35\% on UBIRIS.v2 and 96.75\% on CASIA-Iris-Thousand.

\cite{180} uses an FCN inspired by~\cite{57}. The network starts with a 3$\times$3 kernel that maps the input to the first convolutional layer. This layer consists of 32 channels and uses modified linear units as the activation function. The size of the kernel is kept constant throughout the hidden convolutional layer, while the number of channels and their activation functions remains the same. Finally, in the output layer, the size of the kernel is 3$\times$ 3, but this layer uses an s-type activation function. The performance of this network for the segmentation of frontal iris samples on several public datasets is much smaller in size and complexity than the SPDNN network proposed by~\cite{57} and trained for the off-axis iris sample segmentation task. Due to its lightweight design and high performance for off-axis and frontal iris sample segmentation and processing of input image qualities, the proposed network is well-suited for general deployment in AR or VR devices.

\subsection{Segmentation based on U-shaped neural networks}
The U-Net is a fully convolutional neural network for segmentation first use in medical images. However, benefiting from its particular U-shaped structure, U-Net has shown excellent performance on many segmentation tasks. Therefore, many researchers use U-Net or modified U-shaped networks to perform the segmentation.

The existing methods usually lack the consideration for the geometric constraint that the iris appears only in specific eye regions. Considering the geometric constraint of the iris before image segmentation can improve the segmentation accuracy. \cite{10} regress the bounding box of potential iris regions on top of the original U-Net network to generate an attention mask. This mask is then merged with the discriminative feature map in the model as a weighting function to make the segmentation model more focused on the iris region. The final experimental results demonstrate that considering the geometric constraint that the iris only appears in a specific region of the eye can significantly help to improve the iris segmentation performance.
Meanwhile, many segmentation methods cannot find the actual iris boundary under these non-ideal conditions, resulting in reduced accuracy and reliability of IR. \cite{48} propose a method to more precisely locate the iris image's internal and external iris boundaries. The network model, called Interleaved Residual U-Net, is used for the synthesis and semantic segmentation of iris masks. They also used K-means clustering to select salient points to recover the external boundary of the iris and another set of salient points on the inner side of the mask to recover the internal border. The experimental results show that the proposed iris segmentation algorithm achieves 98.9\% and 97.7\% of the mIoU values for the internal and external boundary estimates, respectively, on the CASIA-Iris-Thousand dataset.

Some studies aim to make the segmentation model smaller and keep the actual segmentation performance, including reducing the network parameters and designing a lightweight network. \cite{46} proposes a novel network model combining U-Net and diluted convolution for iris image segmentation. It replaces all the $3 \times 3$ unpadded convolutions in U-Net with $2 \times 2$ convolutions. Diluted convolution is used instead of the original convolution to extract more global features. The diluted convolution is able to improve segmentation accuracy by better handling image details, and diluted convolution can reduce the computational cost of convolution. \cite{51} proposes an efficient and lightweight U-Net architecture, which is novel in that although this U-shaped network structure consists of 36 layers and uses 148 parameters, its values are an order of magnitude lower than those of other existing networks used for similar applications. The network experiments on images of varying resolution and quality from five standard open-source benchmarks: BioSec, CASIA-V4-Interval, CASIA-V4-Thousand, IITD, and UBIRIS. The experimental results show the network maintains high iris segmentation accuracy and efficiency.

In addition to a single U-Net, \cite{47, 49} combines the advantages of other networks to perform iris segmentation tasks. It can make the network conveniently process complicated iris network. \cite{47} combines Densenet with U-Net and proposes a new network structure that not only reduces the network parameters but also exploits the advantages of U-Net in semantic segmentation. Dense U-Net integrates dense connections into the contraction and expansion paths of U-Net. Compared with traditional U-Net, dense connectivity reduces learning redundancy, enhances information flow, and reduces the required parameters to achieve similar or better performance. Experimental results show that the proposed model can improve accuracy and reduce error rates. In~\cite{49}, an end-to-end encoder-decoder model based on an improved UNet++ is proposed for iris segmentation, referred to as the attention mechanism UNet++. First, the researchers chose EfficientNetV2~\cite{209} as the backbone in~\cite{49} to improve the training speed and reduce the number of network parameters. Secondly, to make full use of the captured location information, enhance the semantic information on the channels to suppress extraneous noise interference, and enhance the discriminability learning of iris regions to improve the recognisability of iris features, the researchers embedded an attention module in the downsampling process of UNet++. Finally, the algorithm uses a pruning scheme to obtain different performance networks, just right for IR in various application scenarios.

\subsection{Segmentation based on other neural networks}
Many of the network structures studied in iris segmentation tasks are based on FCN and U-Net. Some researched network structures are based on classical network structures for iris segmentation.

The residual connection contributes to constructing a deep network structure. In the iris segmentation task, the Property of residual connection can make the network output more precise feature maps. \cite{182} introduces an improved residual connection to extract sufficiently effective iris information. First, this paper uses the dual-attention module to embed the semantic information of the high-level features into the low-level features and to embed the spatial information of the low-level features into the high-level components. The two embedded features are connected to form a practical fusion feature, which achieves accurate localization of the real iris region. \cite{171} establishes residual connections in the convolutional layer between the encoder and decoder structures, thus allowing high-dimensional features to flow in the network and obtaining higher accuracy at the critical layer. In the CASIA-V4-Distance dataset, EER is 0.26\%. In the MICHE-I dataset, the EER is 0.18\%.

% DenseNet
Some studies introduce the dense connection scheme to the iris segmentation network. The dense connection can also optimize the feature extraction process in the segmentation network. \cite{166} proposes IrisDenseNet. The tightly connected IrisDenseNet can determine the true iris boundaries under low-quality images using better information gradient flow between dense blocks. IrisDenseNet is more effective in accurately segmenting difficult regions of the iris region, such as eyelashes and spoof regions. \cite{167} proposes SegDenseNet, which uses four dense blocks to learn the specific iris shape of the model. This algorithm can segment iris patterns from images of eyes before and after cataract surgery. The ASE on the IIITD cataract surgery dataset is 0.98\%.

Similar to U-Net, the hourglass network is a symmetric, fully convolutional structure in which feature mappings are downsampled by pooling operations and upsampled by nearest neighbour upsampling. Some studies use the hourglass network as the backbone for iris segmentation. The~\cite{156} adapted the original hourglass network. At each scale layer, residual connections are applied to the corresponding layers connecting the two sides of the hourglass. This bottom-up and top-down structure is stacked multiple times to capture remote contextual information, allowing accurate segmentation of the iris region by capturing the spatial relationships between the pupil, iris, and edge boundaries.
\cite{156} improves not only segmentation and recognition performance but also reduces the complexity of the model for easy deployment on mobile devices. The smaller \emph{mean normalized Hausdorff distance} (mHdis) indicates a higher detection accuracy, and mHdis reaches a good value in~\cite{156}.
In~\cite{181}, the segmentation network consists of an autoencoder using mainly long-hop and short-hop connections and a superimposed hourglass network between the encoder and decoder. There is continuous up-and-down amplification in the superimposed hourglass network, which helps to extract features at multiple scales. The ASE scores of the proposed method on the UBIRIS.v2, IITD dataset and CASIA-V3-Interval dataset are 0.672\%, 0.916\% and 0.117\%, respectively.

Some typical networks of image segmentation are used to perform the iris segmentation, and they achieve excellent performance on other image segmentation tasks. These models include RefineNet, DeepLab, Mask R-CNN, and SegNet. 
\cite{154} uses RefineNet~\cite{155} to perform iris segmentation task.
RefineNet is a multi-path refinement network that uses a cascade structure with four refinement network units. Each refinement network unit consists of two residual connections. The cascade structure makes the RefineNet can extract precise iris features. Meanwhile, the pre-processing process of this study consists of edge smoothing and Hough transform. However, the segmentation of RefineNet performs satisfactorily only on low-quality datasets. \cite{163} is modified by the DeepLab model~\cite{164}. The DeepLab can obtain more texture details in the iris image than other networks. It simplifies the structure of DeepLab for the iris segmentation task to better match the segmentation task in IR. The experiments were conducted on the visible spectrum iris dataset~\cite{165}. The final experimental segmentation accuracy is 97.5\%. The model of~\cite{160} is based on the Mask R-CNN~\cite{162}. The Mask R-CNN framework enhances the new value of region-based convolution. The average segmentation accuracy of this model on the UBIRIS dataset is 94.8\%. \cite{172} proposes a robust and fast iris segmentation algorithm based on fast R-CNN. \cite{177} is explicitly designed for post-mortem samples, and a SegNet~\cite{170} is used in this approach. The neural network is trained and tested with a completely disjoint set of subjects' iris images. In addition, the mask used for training is annotated and fine-grained to teach the network to localize only the iris regions unaffected by postmortem changes and represent the visible iris texture. For samples collected at ten h postmortem, the EER is less than 1\%.

Using the original iris images as the input of segmentation networks tends to contain some noises. To make the input images of the network contain richer semantic information and ignore irrelevant information, some researchers divide the entire segmentation task into two phases. \cite{159} proposes a CNN-based two-stage iris segmentation method. This iris segmentation method can solve the low-quality noisy images. The first stage includes an undercover filter, noise removal, a Canny edge detector, contrast enhancement, and improved HT to segment the approximate iris boundary. In the second stage, a $21\times 21$ deep CNN detects the real iris boundary. These two stages reduce the processing time and error of iris segmentation. In the MICHE dataset, the ASE is only 0.345\%. \cite{172} utilizes the bounding box found by the fast R-CNN, and the Gaussian mixture model is used to localize the pupil region. The proposed algorithm allows faster segmentation of iris images, which is crucial for real-time IR systems and can even be implemented on mobile devices.

To enable the iris segmentation obtains an extensive application, \cite{173, 174} segment the iris video and iris 3D images, respectively, instead of simple iris images. For volume segmentation of iris videos, \cite{173} proposes the \emph{Flexible Learning-Free Reconstruct of Neural Volumes} (FLoRIN) framework. The FLoRIN framework is a multi-stage pipeline with flexible image processing steps at each stage. The testbed for the framework is a near-infrared iris video dataset in which each subject's pupil rapidly changes size due to visible light stimulation.
FLoRIN can improve the signal of interest features without machine learning. During the segmentation phase, images are loaded into FLoRIN and are selectively processed to improve contrast. Histogram equalisation and Weiner filters incorporate volumetric data into the segmentation process. The images are then thresholded and parametrically thresholded in 2D or 3D using \emph{N-Dimensional Neighborhood Thresholding} (NDNT)~\cite{174}. The neighbourhood size can be specified in 2D or 3D space to merge the volume data into the threshold segmentation. The binarized NDNT output is then passed on to the next stage. Compared to SegNet~\cite{170} and OSIRIS~\cite{175}, FLoRIN achieves a 3.6-fold to an order of magnitude increase in output by merging volumetric information. The matching performance is only slightly reduced. \cite{176} proposes a real-time and accurate 3D eye gaze tracking method for monocular RGB cameras. The key idea is to train a DCNN to automatically extract each eye's iris and pupil pixels from the input image. It combines U-Net~\cite{9} and compressor~\cite{112} to train a compact convolutional neural network suitable for real-time mobile applications. The researchers integrate iris and pupil segmentation and eye closure detection into a 3D eye-tracking framework to achieve real-time 3D eye-tracking results.

\subsection{Summary}
FCN and U-Net are two typical neural network structures used for image segmentation tasks, including iris segmentation. Many researchers have developed their network models based on U-shaped structures. However, these models have yet to demonstrate superior performance compared to other state-of-the-art image segmentation models~\cite{216, 217}. Some studies have attempted to improve the performance of iris segmentation by using a large number of pre-processing operations. However, such methods are computationally complex and may require excessive resources, making them impractical for real-world applications. There is a need to explore more state-of-the-art end-to-end models that can achieve high accuracy with limited computational resources. The methods and results of each study in the specific iris segmentation task are shown in Table.~\ref{tab5}.

\begin{sidewaystable}[htbp]
    \centering
     \caption{Summary of the iris segmentation based on deep learning. \textbf{IIITD} represents IIIT-Delhi Contact lens iris database.}
    \label{tab5}
    \myfontsize 
    \begin{tabular}{cccccccccc}
        \toprule
        \textbf{Year} & \textbf{Reference} &  \textbf{Network} & \textbf{Datasets} & \textbf{Performance} & \textbf{Year} & \textbf{Reference} &  \textbf{Network} & \textbf{Datasets} & \textbf{Performance} \\
        \hline
        \specialrule{0em}{4pt}{4pt}
         2016 & \cite{53} & FCN & \makecell[l]{UBIRIS.v2 \\  CASIA-V4-Distance} & \makecell[l]{EER=0.29\%\\EER=0.20\%} & 2018 & \cite{10} & U-Net& \makecell[l]{UBIRIS.v2\\ CASIA-V4-Distance} & \makecell[l]{MER=0.76\%\\MER=0.38\%} \\
        \specialrule{0em}{4pt}{4pt}
        2017 & \cite{54} & FCN & \makecell[l]{CASIA-Iris-Interval } &\makecell[l]{ACC=90.85\% \\ ACC=99.59\%} & 2018 & \cite{167} & Custom & \makecell[l]{ IIITD Cataract Surgery} & \makecell[l]{ASE=0.98\%} \\
        \specialrule{0em}{4pt}{4pt}
        2017 & \cite{55} & FCN &\makecell[l]{IITD\\Notre Dame\\ CASIA-V4-Interval\\UBIRIS\\CASIA-V5-Aging}  & \makecell[l]{F score=0.8817\\F score=0.9194\\F score=0.8862\\F score=0.2328\\F score=0.8917} & 2018 & \cite{58} & FCN & \makecell[l]{BioSec\\ CASIA-V3\\CASIA-V4-Thousand\\IITD-1\\NICE.I\\CrEye-Iris\\MICHE-I} & \makecell[l]{ASE=0.44\%\\ASE=01.15\%\\ASE=00.61\%\\ASE=01.48\%\\ASE=01.00\%\\ASE=00.96\%\\ASE=00.37\%} \\
        \specialrule{0em}{4pt}{4pt}
        2017 & \cite{56} & FCN & \makecell[l]{CASIA-V5-Aging\\ CASIA-V4-Interval\\IITD} & \makecell[l]{F score=0.925\\F score=0.937\\F score=0.951} & 2018 & \cite{166} &Custom & \makecell[l]{CASIA-V4-Interval\\IITD} & \makecell[l]{ACC=97.10\%\\ACC=98.0\%} \\
        \specialrule{0em}{4pt}{4pt}
        2017 & \cite{159} & Custom & \makecell[l]{MICHE} & \makecell[l]{ASE=0.345\%} & 2018 & \cite{160} & Custom& \makecell[l]{CASIA\\UBIRIS\\ND\\IITD} & \makecell[l]{F score=95.9\\F score=93.2\\F score=95.2\\F score=95.6} \\
        \specialrule{0em}{4pt}{4pt}   
        2017 & \cite{163} & Custom& \makecell[l]{Private} & \makecell[l]{ACC=97.5\%} & 2019 & \cite{46}  &  U-Net & \makecell[l]{CASIA-V4-Interval \\ ND-IRIS-0405 \\ UBIRIS.v2 } & \makecell[l]{ACC=97.36\%\\ ACC=96.74\%\\ ACC=94.81\%} \\
        \specialrule{0em}{4pt}{4pt}
        2018 & \cite{62} & FCN & \makecell[l]{NICE.I } & \makecell[l]{EER=1.13\%} & 2019 & \cite{47} & U-Net& \makecell[l]{CASIA-V4-Interval} & \makecell[l]{ACC=98.36\%} \\
        \specialrule{0em}{4pt}{4pt}
        2018 & \cite{44} & U-Net & \makecell[l]{CASIA } &\makecell[l]{Precision=0.948} & 2019 & \cite{60} & FCN & \makecell[l]{CASIA-V4-Interval \\ IITD \\ UBIRIS.v2 } & \makecell[l]{ACC=99.05\%\\ ACC=98.84\%\\ ACC=99.47\%} \\
        \specialrule{0em}{4pt}{4pt}
        \hline
    \end{tabular}
\end{sidewaystable}

\begin{sidewaystable}[htbp]
    \centering
     \caption{Continued: Summary of the iris segmentation based on deep learning. \textbf{AD} represents after death.}
    \myfontsize 
    \begin{tabular}{cccccccccc}
        \toprule
        \textbf{Year} & \textbf{Reference}  & \textbf{Network} & \textbf{Datasets} & \textbf{Performance} & \textbf{Year} & \textbf{Reference}  & \textbf{Network} & \textbf{Datasets} & \textbf{Performance}\\
        \hline
        \specialrule{0em}{4pt}{4pt}
        2019 & \cite{154}  & Custom & \makecell[l]{CASIA-V4-Interval\\ IITD \\ ND-IRIS-0405} & \makecell[l]{EER=0.213\\EER=0.305\\EER=23.906} & 2020 & \cite{177}  & Custom & \makecell[l]{ Private} & \makecell[l]{(10h AD) EER $<$ 1\%\\(36h AD) EER = 21.45\%} \\
        \specialrule{0em}{4pt}{4pt}
        2019 & \cite{45}  & U-Net & \makecell[l]{CASIA-V4-Interval \\ IITD \\ UBIRIS.v2 \\CASIA-V4-Thousand} & \makecell[l]{F score=0.9842\\F score=0.9764\\F score=0.9699\\F score=0.9785} & 2020 & \cite{180}  & Custom & \makecell[l]{ Bath800\\CASIA-Iris-Thousand\\UBIRIS.v2} & \makecell[l]{ACC=99.13\%\\ACC=99.50\%\\ACC=98.92\%}\\
         \specialrule{0em}{4pt}{4pt}
         2019 & \cite{168}  & Custom & \makecell[l]{  Biosec\\ND-IRIS-0405\\UBIRIS} & \makecell[l]{IoU=87.29\%\\IoU = 89.75\%\\IoU=56.28\%} & 2020 & \cite{181}  & Custom & \makecell[l]{UBIRIS.v2\\IITD \\CASIA-V3-Interval } & \makecell[l]{ASE=0.00672\\ASE=0.00916\\ASE=0.00117} \\
         \specialrule{0em}{4pt}{4pt}
        2019 & \cite{171}  & Custom & \makecell[l]{CASIA-V4-Distance \\ MICHE-I} & \makecell[l]{EER=0.26\%\\EER = 0.18\%} & 2021 & \cite{48}  &U-Net & \makecell[l]{CASIA-Iris-Thousand} & \makecell[l]{mIoU=98.9\%} \\
        \specialrule{0em}{4pt}{4pt}
          2019 & \cite{172}  & Custom & \makecell[l]{ CASIA-Iris-Thousand} & \makecell[l]{ACC= 95.49\%} & 2022 & \cite{49}  & U-Net & \makecell[l]{UBIRIS.v2\\CASIA-V4\\IITD} & \makecell[l]{mIoU=0.9622\\mIoU=0.9774\\mIoU=0.9732} \\
        \specialrule{0em}{4pt}{4pt}
        2019 & \cite{173}  & Custom & \makecell[l]{ Biosec\\ BATH\\ND-IRIS-0405\\UBIRIS\\CASIA-V4-Interval\\IRISSEG-CC\\IRISSEG-EP} & \makecell[l]{EER= 0.074} & 2022 & \cite{51}  & FCN & \makecell[l]{ BioSec\\CASIA-V4-Interval \\  CASIA-V4-Thounsand \\IITD\\ UBIRIS} & \makecell[l]{mIoU=95.44$\pm$0.25\%\\mIoU=97.26$\pm$0.17\%\\mIoU=93.50$\pm$0.23\%\\mIoU=97.09$\pm$0.03\%\\mIoU=92.56$\pm$0.35\%} \\
        \specialrule{0em}{4pt}{4pt}
         2019 & \cite{176} & Custom & \makecell[l]{ Private} & \makecell[l]{EER=8.4241} & 2022 & \cite{182}  & Custom & \makecell[l]{CASIA-Iris-Interval\\ IITD\\ UBIRIS.v2 \\ MICHE.I} & \makecell[l]{EER=0.243\%\\EER=0.0024\%\\EER=2.81\%\\EER=6.08\%}\\
        \specialrule{0em}{4pt}{4pt}
        2020 & \cite{153} & HCNN & \makecell[l]{ CASIA-V3-Interval } &\makecell[l]{ACC=99.5\%} & 2022 & \cite{63}  & FCN & \makecell[l]{ UBIRIS.v2 \\ CASIA-Iris-Thousand} & \makecell[l]{mIoU=95.35\%\\mIoU=96.75\%} \\
        \specialrule{0em}{4pt}{4pt}
        2020 & \cite{156} & Custom & \makecell[l]{CASIA-Iris-M1\\ MICHE-I} & \makecell[l]{mHdis=0.5517\%\\mHdis=1.1107\%}\\
        \specialrule{0em}{4pt}{4pt}
        \hline
    \end{tabular}
\end{sidewaystable}

\section{Presentation Attack Detection}\label{sec5}

\begin{figure}
    \centering
    \includegraphics[width = 1.0\textwidth]{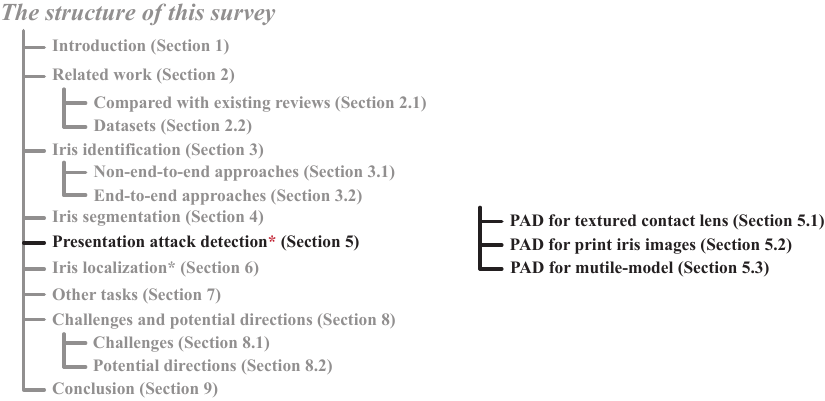}
    \caption{The structure of \textbf{\textit{Presentation Attack Detection}}.}
    \label{structure_a}
\end{figure}

\cite{183} demonstrates that synthetic iris images possess the attack capability for the IR systems by analyzing the quality score distribution of real and iris images synthesized by an adversarial learning-based method. Thus an impostor may subjectively spoof the biometric sensor to circumvent the authentication process. The mechanism that usually protects biometric systems from such attacks is called PAD~\cite{184}. PAD serves as an essential task for IR. Much research has been done in this area. For deep learning-based  PAD in IR, this review discusses the following two aspects: based on typical neural networks and based on some novel neural networks. The structure of this section is shown in Fig.~\ref{structure_a}.

\subsection{Presentation attack detection for textured contact lens}

AlexNet and DenseNet, two networks commonly used in image analysis tasks, performed the PAD for the TCLs. In~\cite{190}, the pre-trained AlexNet and SVM were used for feature extraction and classification of texture patterns of iris images. The results show that the total error of this PAD algorithm is 10.21\%. Deep learning can be inferred to classify real and attacked iris images based on discriminative information encoded by the features. Further analysis revealed that \emph{Attack Presentation Classification Error Rate} (APCER) = 11.79\% and \emph{Bonafide Presentation Classification Error Rate} (BPCER) = 2.28\%. \cite{7} creates a combined feature set using features extracted from the Dense Connected Contact Lens Classification Network (DCCNet) based on DenseNet-121 and manually extracted features. Then, top-k features are selected using the Friedman test to determine the best features and fused by fraction level. Experimental results show that this approach improves the discrimination of eye image attacks on the iris system of people wearing TCLs.

Image block and edge enhancement have also been applied to PAD tasks for TCLs. The network of~\cite{129} samples the input iris image, splits the original 512$times$64 pixel images into 32$times$32 pixel image blocks, and then allows each image block to be rotated at four different angles. These non-overlapping image blocks are finally used to complete the classification decision. In~\cite{199}, the input to the model consists of both original and edge-enhanced images, and the combination of original and edge-enhanced iris images in the pre-processing stage by manual techniques enables the CNN to learn more effective discriminative features for binary classification. Experimental results demonstrate the model's effectiveness on eye images of contact lens wearers from the LivDet-2017 and IIITD contact lens datasets for the PAD task.

\cite{191} explores the effects of gender on the PAD system. This paper uses three different methods for the experiments. The first is LBP~\cite{192} for feature extraction. The second uses pre-trained VGG-16 and PCA in the feature extraction step. Both of the above methods use SVM for classification. The third method uses the computationally efficient network MobileNetV3-Small~\cite{193} to train from scratch. The results of this paper in the Notre Dame Contact Lens Database-2013 dataset show that PAD is significantly less protective of female users than male users.

\subsection{Presentation attack detection for print iris images}
\cite{195} explores the possibility of using EMS-based features to distinguish between real and deceptive eyes. In the design of the network structure, it replaces the two-dimensional convolution of ResNet-18~\cite{204} with a one-dimensional one and then performs a PAD task against iris-printed images. The technique has a high average classification rate across different attack scenarios and requires only 1.5 seconds of eye movement data to make a deceptive detection decision. The model achieves an EER of 3.62\% on the ETPAD-v2 dataset~\cite{205}.

\subsection{Presentation attack detection for mutile-model}
The well-known VGGNet and MobileNet are used in iris PAD tasks. \cite{40} uses an pre-trained VGG-16~\cite{106} and a scratch-trained MobileNet~\cite{193} for the experiments. In the feature extraction phase, it extracts features in the last four convolutional layers of the pre-trained and scratch-trained networks, respectively. PCA is then used to reduce the dimensionality of the features extracted from each layer. In the classification phase, SVMs are trained to make the PAD decisions for each layer, and logistic regression is used to fuse the decision scores of each classifier to produce the final PAD decisions. The \emph {Average Classification Error Rate} (ACER) for the multilayer solution based on the pre-trained VGG-16 from the University of Notre Dame is 2.31\%. In the Notre Dame dataset, the MobileNet-based model achieves an ACER of 8.89\%. In addition to the above networks, \cite{200} uses an automatic encoder based on supervised feature learning methods to perform iris PAD tasks, including printed iris images, iris images of contact lens wearers and synthetic iris images. This structure is able to use class information in the training phase to reduce reconstruction and classification errors. Experimental results show that the proposed method generalises well between experiments with different datasets.

In multiple attacks of PAD, fusing local and global features helps the model extract discriminative features of true and false iris images. \cite{39} uses CNN for feature extraction of local and global eye regions and then uses SVM for the classification of these features. In the Warsaw-2017 dataset and the Notre Dame Contact Lens Database-2015 dataset, the ACER is 0.016 \% and 0.292 \%, respectively. APCER is the rate at which the attacked image is incorrectly classified as a true sample. BPCER is the proportion of real images that are incorrectly classified as attack samples. The average of the APCER and BPCER is ACER. \cite{185} combines local and global iris features in a multilevel redundant discrete wavelet transform domain with features learned by VGG-16. Texture variations between the real iris image and the attack iris image are encoded. They are then reduced to low-dimensional vectors by principal component analysis. Finally, MLP is used for classification. The algorithm Is tested on combined large datasets~\cite{185} consisting of LivDet-2013-Warsaw, Notre Dame Contact Lens Database-2013, Notre Dame Contact Lens Database-2015, and the joint spoofing database~\cite{208}, which contains over 270,000 real and attack iris images, with a total error of 1.01\%.

In contrast, to feature fusion, \cite{143, 194, 196} use other methods to improve the PAD model's feature extraction ability or classification ability. \cite{194} uses the YOLO to localize the RoI in pre-processing. Iris texture features in the RoI are then extracted using a variety of manual and CNN-based methods. The best k-features are then identified as the best feature set by a Friedman test, and global classification is performed using SVM-based fraction-level fusion. \cite{143} uses a sliding window to decompose the normalized iris image into 28 patches. The CNN then outputs the classification results for each iris patch. The proposed method approaches 100\% accuracy on the ND-Contact and CAISA-Iris-Fake datasets. \cite{196} chooses a Euclidean loss function to train this network rather than the traditional softmax loss, as the former can directly calculate confidence scores rather than just generating class labels. The final experimental results show that the model generalizes well across datasets and across sensors.

\subsection{Summary}
In the PAD task, many researchers are making the biometric system protected from attacks by improving the accuracy of the IR task. The general attacks are printed iris images, contact lenses, and synthetically generated iris images. The PAD task of IR also has some unique methods to handle it, such as eye movement analysis, data enhancement, and multimodality. The experimental results obtained can be found in Table.~\ref{tab6}.

\begin{sidewaystable}[htbp]
    \centering
     \caption{Summary of the iris PAD based on deep learning. \textbf{Tr} represents train dataset. \textbf{Te} represents test dataset. \textbf{LivDetW15} represents LivDet-Iris-2015-Warsaw. \textbf{TCL} represents the eye images wearing textured contact lenses. \textbf{ETPAD.v2} represents Eye Tracker Print-Attack Database. \textbf{IIITD} represents the IIIT-Delhi Contact lens iris database. \textbf{RI} represents Recognition of Iris.}
    \label{tab6}
    \myfontsize 
    \begin{tabular}{cccccccc}
        \toprule
        \multirow{2}*{\textbf{Year}} & \multirow{2}*{\textbf{Reference}} & \multirow{2}*{\textbf{Network}} & \multirow{2}*{\textbf{Datasets}} & \multicolumn{4}{c}{\textbf{Performance}} \\
        \cline{5-8} 
        ~ & ~ & ~ & ~ & \textbf{Print} & \textbf{Plastic} & \textbf{TCL} & \textbf{Kindle} \\
        \hline
        \specialrule{0em}{4pt}{4pt}
         2018 & \cite{184} & VGG & \makecell[l]{LivDetW15\\Automated Targeting and RI\\Train: Automated Targeting and RI Test: LivDetW15 \\ Train: LivDetW15 Test: Automated Targeting and RI} & \makecell[l]{HTER=0.075\%\\HTER=0.000\%\\HTER=0.312\% \\HTER=49.99\%} & \makecell[c]{- \\ - \\ - \\ 
         -} & \makecell[c]{- \\ - \\ - \\ 
         -} & \makecell[c]{- \\ - \\ - \\ 
         -}\\
        \specialrule{0em}{4pt}{4pt}
         2018 & \cite{185} & VGG-16 & Combined large datasets & \multicolumn{4}{l}{\textit{(without pointout attack class)} Total Error=1.01\%} \\ 
        \specialrule{0em}{4pt}{4pt}
         2020 & \cite{40} &  MobileNet & \makecell[l]{Notre Dame \\ IITD-WVU} & \multicolumn{4}{c}{\textit{(without pointout attack class)} \makecell[l]{HTER = 2.31\% \\ HTER = 15.0\%}} \\
         
        \specialrule{0em}{4pt}{4pt}
         2018 & \cite{190} & AlexNet & Private & - & - & \makecell[l]{Total Error=10.21\%} & - \\
        \specialrule{0em}{4pt}{4pt}
        2020 & \cite{189} & DenseNet & \makecell[l]{Private \\ LivDet-2017} & \multicolumn{4}{l}{\textit{(without pointout attack class)} \makecell[l]{TDR=98.58\% \\ APCER = 1.3\% }} \\
        \specialrule{0em}{4pt}{4pt}
           2021 & \cite{194} & Custom & \makecell[l]{IIITD CLD\\ND CLD\\ND-LivDet\\IIITD-CSD\\Clarkson2015\\Clarkson2017} & \multicolumn{4}{l}{\textit{(without pointout attack class)} \makecell[l]{EER=1.07\%\\EER=2.02\%\\EER=2.92\%\\EER=1.62\%\\EER=3.85\%\\EER=3.25\%}} \\
        \specialrule{0em}{4pt}{4pt}
             2022 & \cite{195} &ResNet-18 & \makecell[l]{ETPAD.v2} & \makecell[l]{EER=3.62\%} & - & - & - \\
        \specialrule{0em}{4pt}{4pt}
         2020 & \cite{50} & Custom & \makecell[l]{NDIris3D} & \makecell[l]{EER=4.39\%} & - & - & - \\
        \specialrule{0em}{4pt}{4pt}
        \hline
    \end{tabular}
\end{sidewaystable}

\begin{sidewaystable}[htbp]
    \centering
     \caption{Continued: Summary of the iris PAD based on deep learning. \textbf{Tr} represents train dataset. \textbf{Te} represents test dataset. \textbf{LivDetW15} represents LivDet-Iris-2015-Warsaw. \textbf{LivDetC17} represents LivDet-Iris-2017-Clarkson. \textbf{ND13-\uppercase\expandafter{\romannumeral1}} represents Notre Dame Contact Lens 2013-I. \textbf{ND13-\uppercase\expandafter{\romannumeral2}} represents Notre Dame Contact Lens 2013-II. \textbf{CE} represents contraction-expansion CNN. In~\cite{197}, 1, 2, and 3 correspond to F-CNN, S-CNN, and I-CNN trained on the IrisID dataset, while 4 and 5 correspond to S-CNN and F-CNN trained on LivDet-Iris-2015-Warsaw. \textbf{TCL} represents the eye images wearing textured contact lenses. \textbf{IIITD} represents the IIIT-Delhi Contact lens iris database. \textbf{ND} represents Notre Dame cosmetic contact lens database 2013.}
    \myfontsize 
    \begin{tabular}{cccccccc}
        \toprule
        \multirow{2}*{\textbf{Year}} & \multirow{2}*{\textbf{Reference}} & \multirow{2}*{\textbf{Network}} & \multirow{2}*{\textbf{Datasets}} & \multicolumn{4}{c}{\textbf{Performance}} \\
        \cline{5-8} 
        ~ & ~ & ~ & ~ & \textbf{Print} & \textbf{Plastic} & \textbf{TCL} & \textbf{Kindle} \\
        \hline
         \specialrule{0em}{4pt}{4pt}
         2018 & \cite{196} & Custom & \makecell[l]{Train: BioEye Research Consortium-Iris-Face Test: LivDetW15 \\Train: BioEye Research Consortium-Iris-Face Test: CASIA-IF} & \makecell[l]{CC = 95.1\% \\ CC = 100\%} & \makecell[c]{- \\ \makecell[l]{43.75\%}} & \makecell[c]{- \\ \makecell[l]{9.30\%}} &\makecell[c]{- \\ - }\\
        \specialrule{0em}{4pt}{4pt}
          2018 & \cite{203} & Custom & \makecell[l]{ND-Contact\\ CASIA-Iris-Interval \&Syn} & \multicolumn{4}{c}{\textit{(without pointout attack class)} \makecell[l]{CCR= 99.58\% \\ CCR= 100\% }}\\
        \specialrule{0em}{4pt}{4pt}
          \multirow{7}*{2019} & \multirow{7}*{\cite{197}}  & $1 + 2$ & \makecell[l]{IrisID \\ LivDetW15 \\ BioEye Research Consortium-IF} &  \makecell[l]{TDR = 99.18\% \\ TDR = 99.87\% \\ TDR = 100\%} & \makecell[l]{100\% \\ - \\ 100\%} & \makecell[l]{- \\ - \\ 100\%} & \makecell[l]{100\% \\ - \\ -} \\
        \specialrule{0em}{4pt}{4pt}
        ~ & ~ & $1 + 2 + 3$ & \makecell[l]{IrisID \\ LivDetW15 \\ BioEye Research Consortium-IF} & \makecell[l]{TDR = 99.73\% \\ TDR = 99.82\% \\ TDR = 100\%} & \makecell[l]{100\% \\ - \\ 100\%} & \makecell[l]{- \\ - \\ 100\%} & \makecell[l]{100\% \\ - \\ -} \\
        \specialrule{0em}{4pt}{4pt}
        ~ & ~ & $1 + 2 + 4$ & \makecell[l]{IrisID \\ LivDetW15 \\ BioEye Research Consortium-IF} & \makecell[l]{TDR = 99.18\& \\ TDR = 100\% \\ TDR = 100\%} & \makecell[l]{100\% \\ - \\ 100\%} & \makecell[l]{- \\ - \\ 100\%} & \makecell[l]{100\% \\ - \\ -} \\
        \specialrule{0em}{4pt}{4pt}\
        2020& \cite{198} & Custom & \makecell[l]{IIIT-WVU} & \multicolumn{4}{c}{ \textit{(\textbf{Print}} and \textbf{\textit{TCL}}) ACER=26.19\%} \\
        \specialrule{0em}{4pt}{4pt}
        2020 & \cite{7} & Custom & \makecell[l]{ND13-\uppercase\expandafter{\romannumeral2} } & - & - & \makecell[l]{EER=0.2\%} & - \\
         \specialrule{0em}{4pt}{4pt}
         2017 & \cite{129} & Custom & \makecell[l]{IIITD \\ND}  & - & - & \makecell[l]{CCR=94.65\% \\CCR=92.60\%} & - \\
           \specialrule{0em}{4pt}{4pt}
        \hline
    \end{tabular}
\end{sidewaystable}

\begin{sidewaystable}[htbp]
    \centering
     \caption{Continued: Summary of the iris PAD based on deep learning. \textbf{WBPv1} represents Warsaw-BioBase-PostMortem-Iris-v1. \textbf{LivDetW13}LivDet-Iris-2013-Warsaw. \textbf{TCL} represents the eye images wearing textured contact lenses. \textbf{Combined Spoofing Database} represents the combined dataset, including Synthetic, Multi-sensor, IIIT-Delhi Contact Lens, IIITD Iris Spoofing, and IIT Delhi Iris.}
    \myfontsize 
    \begin{tabular}{ccclcccc}
        \toprule
        \multirow{2}*{\textbf{Year}} & \multirow{2}*{\textbf{Reference}} & \multirow{2}*{\textbf{Network}} & \multirow{2}*{\textbf{Datasets}} & \multicolumn{4}{c}{\textbf{Performance}} \\
        \cline{5-8} 
        ~ & ~ & ~ & ~ & \textbf{Print} & \textbf{Plastic} & \textbf{TCL} & \textbf{Kindle} \\
        \hline
        \specialrule{0em}{4pt}{4pt}
         2022 & \cite{199} & Custom & \makecell[l]{Train: ND-PSID Test: UnMIPA\\ Train:MUIPA Test: UnMIPA} & \makecell[c]{-} & \makecell[c]{- \\ -} & \makecell[l]{ACER=20.1\%\\ACER=10.3\%} & \makecell[c]{- \\ -} \\
         \specialrule{0em}{4pt}{4pt}
        \multirow{5}{*}{2021} & \multirow{5}{*}{\cite{191}} & MobileNetV3-Small & Notre Dame Contact Lens Database-2013 & \makecell[c]{- \\ - \\ -} & \makecell[c]{- \\ - \\ -} & \makecell[l]{\textit{(Male)} HTER = 4.16\% \\ \textit{(Female)} HTER =4.72\% \\ \textit{(Both)} HTER = 4.30\% } & \makecell[c]{- \\ - \\ -} \\
        \specialrule{0em}{4pt}{4pt}
        ~ & ~ & VGG-16 & Notre Dame Contact Lens Database-2013 & \makecell[c]{- \\ - \\ -} & \makecell[c]{- \\ - \\ -} & \makecell[l]{\textit{(Male)} HTER = 6.22\% \\ \textit{(Female)} HTER = 8.08\% \\ \textit{(Both)} HTER = 6.7\% } & \makecell[c]{- \\ - \\ -}  \\
        \specialrule{0em}{4pt}{4pt}
          2022 & \cite{200} & Custom & \makecell[l]{Combined Spoofing Database} & \multicolumn{4}{l}{\textit{(without pointout attack class)} ACC = 87.58\%} \\
        \specialrule{0em}{4pt}{4pt}
           \multirow{8}{*}{2022} & \multirow{8}{*}{\cite{201}} & ResNet-50 & \makecell[l]{Clarkson \\ Notre Dame \\ IIITD-WVU} & \multicolumn{4}{c}{ \textit{(\textbf{Print}} and \textbf{\textit{TCL}}) \makecell[l]{ACER = 5.12\% \\ ACER = 7.47\% \\ ACER = 10.05\%}} \\
        \specialrule{0em}{4pt}{4pt}
        ~ & ~ & VGG-16 & \makecell[l]{Clarkson \\ Notre Dame \\ IIITD-WVU} & \multicolumn{4}{c}{ \textit{(\textbf{Print}} and \textbf{\textit{TCL}}) \makecell[l]{ACER = 5.81\% \\ ACER = 11.64\% \\ ACER = 18.53\%}} \\
        \specialrule{0em}{4pt}{4pt}
        ~ & ~ & MonileNet & \makecell[l]{Clarkson \\ Notre Dame \\ IIITD-WVU} & \multicolumn{4}{c}{ \textit{(\textbf{Print}} and \textbf{\textit{TCL}}) \makecell[l]{ACER = 0.84\% \\ ACER = 10.28\% \\ ACER = 10.79\%}} \\
            2018 & \cite{39} & Custom & \makecell[l]{Warsaw-2017 \\ Notre Dame Contact Lens Database-2015} & \makecell[c]{\makecell[l]{ACER = 0.016\%} \\ - } & \makecell[c]{- \\ -} & \makecell[c]{- \\ \makecell[l]{ACER = 0.292\%}} & \makecell[c]{ - \\ -} \\
        \specialrule{0em}{4pt}{4pt}
         2016 & \cite{143} & Custom & \makecell[l]{ND-Contact \\ LivDetW13} & \makecell[c]{-  \\ \makecell[l]{- \\ CCR = 100\%}} & \makecell[c]{- \\ -} & \makecell[c]{\makecell[l]{CCR = 98.15\%} \\ -} & \makecell[c]{- \\ -} \\
        \specialrule{0em}{4pt}{4pt}
        \hline
    \end{tabular}
\end{sidewaystable}

\section{Iris Localization}\label{sec6}
In the localization task of IR, the YOLO model has a crucial role in the localization of iris~\cite{122,123,145}. It is mainly due to the advantages of the YOLO model, such as fast speed and high generalization ability. Moreover, it is more suitable for IR, and there is rarely overlap.

\cite{122} model based on YOLO.v2~\cite{123} predicts the image's bounding box and class probabilities directly by a cost-effective deep learning neural network compared to previous target detection work. The YOLO.v2 model can effectively reduce the computation time. The design is more efficient as it does not require iris and sclera segmentation computation. The model jointly labels the visible eye images as partial iris and sclera regions and trains an identity classifier to infer the correct individual identity. The proposed method achieves \emph{mean Average Precision} (mAP) of 99\% for the joint recognition of the iris and sclera.
The multi-task convolutional neural network learning method proposed by~\cite{203} can perform iris localization and PAD, or the iris localization task can be performed separately. The process is based on YOLO.v2, which directly regresses the parameters of the iris bounding box and calculates the probability of presenting an attack from the input eye image. The detection accuracy of the method is 100\% in LivDet-Iris-2015-Warsaw and 99.83\% in CASIA-Iris-Fake.
\cite{145} compares two window-based detectors with the Daugman iris localization method. The first is a histogram of oriented gradients and a linear SVM classifier based on a histogram of oriented gradient features. The second one is a deep learning-based YOLO target detector with fine-tuning of the detector. Experimental results show that the YOLO target detector-based detector outperforms the other detectors in terms of accuracy and runtime. It uses multiple datasets for the experiments of the three methods. Table.~\ref{tab7} shows a dataset where the YOLO model has the highest ACC. The YOLO model has a higher ACC than the other two methods on all the datasets where the experiments were conducted.
\cite{194} uses a YOLO model to localize the RoI, where the YOLO model is tightly integrated with the iris region without any pattern loss while preserving important texture details. The aim is to improve the discrimination between the real iris and attack samples and the performance of iris PAD.

\cite{146} proposes to use Iris R-CNN for iris segmentation and localization. The Iris R-CNN seamlessly integrates segmentation and localization into a unified framework to generate the normalized iris images required for IR. \cite{146} proposes a two-circle region suggestion network and a two-circle classification and regression network to efficiently capture the pupil and iris circles and improve iris localization accuracy. In addition, this study proposes a region of interest normalization scheme to achieve a new ensemble operation on bicircular regions. The performance of the iris R-CNN experiment on three sub-datasets of the MICHE dataset for iris localization was good. The performance is measured using Hausdorff distance to calculate the shape similarity between the actual inner and outer ring boundaries and the estimated boundaries. Smaller HALoc values mean higher localization accuracy.

The specific performance of the above studies on iris localization tasks is shown in Table.~\ref{tab7}.

\begin{table}[htbp]
    \centering
     \caption{Summary of the iris localization tasks based on deep learning. \textbf{IP5} represents iPhone5. \textbf{GS4} represents Samung Galaxy S4. \textbf{GT2} represents Samsung Galaxy Tab2. \textbf{PSNR} represents peak signal-to-noise ratio. \textbf{SSIM} represents structural similarity index measure. \textbf{NTHU-DDD} represents National Tsing Hua University Drowsy Driver Detection Video Dataset. }
   \label{tab7}
    \myfontsize 
    \begin{tabular}{cccccc}
        \toprule
        \textbf{Year} & \textbf{Reference} & \textbf{Task} & \textbf{Network} & \textbf{Datasets} & \textbf{Performance}\\
        \hline
         \specialrule{0em}{4pt}{4pt}
         2018 & \cite{203} & localization & \makecell[l]{YOLO.v2} & \makecell[l]{ND-Contact\\ CASIA-Iris-Interval \&Syn}& \makecell[l]{CCR= 99.58\% \\ CCR= 100\% } \\
         \specialrule{0em}{4pt}{4pt}
        2018 & \cite{145} & localization& \makecell[l]{YOLO} & \makecell[l]{\cite{206}} & \makecell[l]{ACC=98.72\%\\ACC=98.48\%\\ACC=99.71\%} \\
        \specialrule{0em}{4pt}{4pt}
        2021 & \cite{122} & localization & \makecell[l]{YOLO.v2}& \makecell[l]{self-made} & \makecell[l]{mAP=99\%.} \\
        \specialrule{0em}{4pt}{4pt}
        2021 & \cite{194} & localization & \makecell[l]{YOLO} & \makecell[l]{IIITD CLD\\ND CLD\\ND-LivDet\\IIITD-CSD\\Clarkson2015\\Clarkson2017}& \makecell[l]{EER=1.07\%\\EER=2.02\%\\EER=2.92\%\\EER=1.62\%\\EER=3.85\%\\EER=3.25\%}\\
        \specialrule{0em}{4pt}{4pt}
         2022 & \cite{146} & localization & \makecell[l]{R-CNN} & \makecell[l]{MICHE IP5\\ MICHE GS4\\MICHE GT2} & \makecell[l]{HALoc=1.18\\HALoc=1.43\\HALoc=1.67} \\
             \specialrule{0em}{4pt}{4pt}
        \hline
    \end{tabular}
\end{table}

\section{Other tasks}\label{sec7}
In the process of IR, deep learning techniques are used not only for iris identification, iris segmentation, iris PAD, and iris localization but also for other tasks, such as image enhancement and fatigue driving detection. Deep learning techniques also play a significant role in these tasks.

Researchers sometimes acquire low-resolution iris images in IR due to their relaxed acquisition conditions. However, low-resolution images tend to compromise the performance of the model. Thus, various super-resolution techniques for iris images are emerging, especially using CNNs.\cite{149} uses three different super-resolution CNNs to improve IR performance, namely SRCNN~\cite{150}, VDCNN~\cite{151}, and SRGAN~\cite{152}.SRCNN is one of the image super-resolution and one of the first proposed CNN architectures in the field. vDCNN relies on a deep CNN inspired by VGG. sRGAN relies on two different CNNs and a new objective function scheme to recover finer texture details from low-resolution images.
\cite{207} combines the super-resolution network of GAN~\cite{148} with the iris feature extraction network, which helps to retain identity information. Inspired by SRGAN~\cite{152}, \cite{207} introduces the concept of adversarial into the triplet network and proposes a new super-resolution structure. Meanwhile, replacing the traditional cross-entropy function with triplet loss facilitates the recognition of the generated super-resolution iris images.
\cite{147} proposes \emph{Conditional Generative Adversarial Network} (cGAN)-based iris image augmentation techniques to improve recognition performance. The technique uses only the iris region as the input to the cGAN model, and experimental results demonstrate that the recognition performance is improved.

\cite{70} uses a DNN estimated by the degree of eye openness to identify fatigued drivers. It uses a segmentation network with a lightweight U-Net structure to segment the eye image and accurately extract pupil and iris features from the video image. The extracted feature maps are then used to guide the decision network to estimate the openness of the eye. The detection method tests using the National Tsing Hua University drowsy driver detection video dataset, with a fatigue detection accuracy of 96.72\%.

The specific performance of the above studies is shown in Tab.~\ref{tab8}.

\begin{table}[htbp]
    \centering
     \caption{Summary of the other tasks based on deep learning. \textbf{PSNR} represents peak signal-to-noise ratio. \textbf{SSIM} represents structural similarity index measure. \textbf{NTHU-DDD} represents National Tsing Hua University Drowsy Driver Detection Video Dataset. }
   \label{tab8}
    \myfontsize 
    \begin{tabular}{cccccc}
        \toprule
        \textbf{Year} & \textbf{Reference} & \textbf{Task} & \textbf{Network} & \textbf{Datasets} & \textbf{Performance}\\
        \hline
         \specialrule{0em}{4pt}{4pt}
        2018 & \cite{149} & Image enhancement & \makecell[c]{VDDCNN \\ SRGAN} & \makecell[l]{VSSIRIS} & \makecell[l]{PSNR = 25.26 \\ PSNR = 24.59 } \\
        \specialrule{0em}{4pt}{4pt}
          2018 & \cite{207} & Image enhancement & Custom & \makecell[l]{CASIA} & \makecell[l]{PSNR = 30.97} \\
        \specialrule{0em}{4pt}{4pt}
          2019 & \cite{147} &Image enhancement& cGAN & \makecell[l]{MICHE} & \makecell[l]{EER = 12.79\%} \\
        \specialrule{0em}{4pt}{4pt}
             2020 & \cite{70} & Fatigue driving detection & Custom & \makecell[l]{NTHU-DDD} & \makecell[l]{Precision=96.72\% \\Recall Rate=93.63\%\\ F1=95.16} \\
        \specialrule{0em}{4pt}{4pt}
        \hline
    \end{tabular}
\end{table}

\section{Challenges and potential directions}\label{sec8}
Existing IR methods already have good recognition performance with relatively stable and secure iris features, but the IR recognition process still faces a number of challenges. This section summarises the existing IR research and analyses these challenges. Meanwhile, neural network techniques, represented by deep learning, have become increasingly popular in recent years. In this section, we summarise potential directions for deep learning-based IR based on this review, which will be helpful for researchers on IR tasks.

\subsection{Challenges}
\subsubsection{Absence of generalisation methods}
The application of deep learning techniques to IR has greatly improved its performance of IR. Many different network frameworks have been used for IR, the drawback of which is that the performance of the proposed method is very often dependent on the iris dataset chosen. Applying it to other iris datasets is likely to fail. The challenge with deep learning is to find a suitable learning framework and choose the right network parameters for the task in question. Despite the advantages of deep learners, there are still issues such as time complexity and computational intensity in deep learning, and in particular, the question of how to lighten the network is also an urgent issue to be addressed.

\subsubsection{Unconstrained imaging environments}
There are also many iris images that are acquired under less constrained conditions, such as lighting conditions, imaging angles, shooting distances, and glasses, which are unavoidable under realistic conditions. There is a large amount of noise in the images. In such cases, it is particularly important to reasonably design a DNN for robust and accurate recognition of the iris.

\subsection{Potential directions}

\subsubsection{Feature extraction based on unsegmention }
Most IR tasks require prior segmentation of the iris image to extract features. The segmented iris image enables DNNs to extract iris features more accurately, but the segmentation operation introduces additional computation to the recognition process. Even though some studies use unsegmented iris images, they have yet to demonstrate that their performance is comparable to those obtained with segmented iris images. Hence, there is a need to develop feature extraction methods that can accurately capture iris features without segmenting the iris image and without degradation in performance. This will simplify the IR process and improve the inference speed of IR systems.

\subsubsection{Iris video capture and analysis}
Only a few studies have considered eye video. For instance, \cite{195} for detecting iris print attacks by analyzing the captured eye movement signals with deep learning models. Studies of eye video not only allow multiple iris images to be acquired and protect the system from, for example, printed images. In many cases, iris biometric capture devices acquire multiple images in sequence and combine each image's segmentation and matching results to enhance the results. So far, segmentation operates on each image individually. However, this approach does not exploit the potential performance gains from treating consecutive images as volumetric data~\cite{173, 176}. 

\subsubsection{Binocular collaboration}
Each individual's iris texture information is unique, as are identical twins. Moreover, the iris textures of the left and right eyes are inconsistent for the same person. \cite{33} also classifies left and right iris images. However, the formation of the iris is determined by genetic inheritance, and human gene expression determines the iris's morphology, physiology, colour, and general appearance. Since both the left and right iris originates from the same gene type, the two iris cannot be completely unrelated.
On the one hand, the existing IR studies lack exploring left-right iris correlation. On the other hand, it may be difficult for existing deep learning models to explore these potential iris features if the correlation between the irises of the two eyes of the same person is demonstrated and can play a role in the IR process. Then the number of iris images collected would be reduced by half, significantly reducing the workload.

\subsubsection{Federal Learning}
The current state of IR research often prioritizes the pursuit of improved performance, neglecting the importance of protecting iris feature samples. This oversight can have significant consequences, including the potential compromise of individuals' privacy and security. Therefore, it is critical to recognize the importance of implementing robust security measures to protect biometric data and feature samples. By doing so, researchers can ensure that advancements in IR technology are made with privacy and security in mind and that individuals' rights to privacy are respected. Federated learning is a distributed machine learning approach that aims to create more accurate and robust global models by merging models trained on multiple devices. Unlike traditional centralized deep learning methods, federated learning does not require sending data to a central server, providing a privacy-preserving alternative for deep learning-based IR. 

\subsubsection{Model lightweighting}

Most existing high-performance iris recognition systems rely on very deep network layers with a large number of training parameters. However, in real-world application scenarios, iris recognition algorithms are often deployed on small mobile devices rather than the high-performance servers used to train the networks. These mobile devices are often limited in computational performance and memory, which poses a significant challenge to deploying networks with deep layers that perform well on these devices. This limitation on the size of deep networks makes it difficult to achieve optimal performance in iris recognition on mobile devices. In order to meet the deployment requirements of IR systems, it is advisable to employ network pruning~\cite{214} and knowledge distillation~\cite{215} techniques alongside the development of lightweight network models. Network pruning is a widely adopted technique in deep learning for reducing the number of parameters in a model. The main objective of this technique is to reduce the computational and memory overheads of a DNN by removing redundant weights and connections in the network, thus shrinking its size. This can help improve the efficiency and speed of the network without compromising its performance. Knowledge distillation is a popular technique used in deep learning for transferring the knowledge of a large, complex model, known as the teacher model, to a smaller and simpler model, known as the student model. The core idea behind knowledge distillation is to train the student model to approximate the behaviour of the teacher model by minimizing the difference between their predictions. By adopting these methods, it is possible to reduce the computational and memory overheads of the network while maintaining its performance. The use of these techniques enables the creation of efficient and faster IR systems that can be deployed on low-end devices such as mobile phones or IoT devices. Therefore, it is essential to consider the application of these techniques during the design and development of IR systems for real-world scenarios.

\subsubsection{Application of advanced model}
Numerous novel models have been developed for visual tasks in recent years, achieving impressive performance with distinct advantages~\cite{210, 211, 212, 213}. One such model is ViT, which is a visual model that has been developed based on the Transformer language model. Despite its lightweight structure, ViT is able to achieve comparable performance to DNNs. Currently, the ViT architecture is the foundation for many of the most advanced models in computer vision. However, Our survey indicates that the most advanced models in visual tasks have yet to be utilized in IR. Swin Transformer~\cite{210} adopts a hierarchical local attention mechanism that adaptively adjusts each image token's size, resulting in improved scalability and computational efficiency when processing large-scale image data compared to ViT. MobileViT is a lightweight Transformer network for mobile devices that uses separable convolution to reduce the amount of computation and the number of parameters. LeViT combines the advantages of Transformer and convolutional neural networks by embedding Transformer's attention mechanism in existing convolutional neural networks, thus achieving faster inference speeds. The application of these state-of-the-art models can drive the development of IR.

\section{Conclusion}\label{sec9}
This paper provides a comprehensive review of deep learning-based IR, bridging the lack of a comprehensive review related to deep learning in the field of IR. A total of 120 papers have been collected to support this work. The purpose of this paper is to discuss deep learning-based IR tasks. The paper begins with some information related to IR, including an analysis of IR background in Section.~\ref{sec1} and related reviews of an introduction to commonly used public datasets in Section.~\ref{sec2}. The survey reviews sections immediately follow with the most important work of this paper. In Section.~\ref{sec3}, we discuss the recognition task of deep learning in IR, and the process of the identification task can be divided into non-end-to-end and end-to-end processes. In the non-end-to-end process, we discuss it according to the process, including pre-processing, feature extraction and matching. In the end-to-end process, we discuss it in terms of typical and novel neural networks, respectively. U-Net and FCN show their excellent performance on the segmentation task. In addition, the PAD task on IR is also studied.
Moreover, the YOLO model achieves good results on the iris localization task. GAN performs well in iris image enhancement. There are some other tasks in which deep learning also plays an important role. All these studies achieved excellent performance. This survey concludes with a summary of some typical challenges of IR and suggests potential directions for IR. They can inspire future research in the field of IR.

\bibliography{sn_bibliography}

\end{document}